\documentclass[11pt]{article}

\usepackage[final]{acl}

\usepackage{times}
\usepackage{latexsym}

\usepackage[T1]{fontenc}

\usepackage[utf8]{inputenc}
\usepackage{amsfonts}

\usepackage{microtype}

\usepackage{inconsolata}

\usepackage{graphicx}
\usepackage{algorithm}

\usepackage{algorithmicx}
\usepackage{algpseudocode}
\usepackage{amsmath}
\usepackage{booktabs}
\usepackage{multirow}
\title{SAGE: Multi-Agent Self-Evolution for LLM Reasoning}

\author{Yulin Peng \textsuperscript{\rm 1}, Xinxin Zhu \textsuperscript{\rm 1, \rm 2}, Chenxing Wei \textsuperscript{\rm 1, \rm 2}, Nianbo Zeng \textsuperscript{\rm 1, \rm 2}, Leilei Wang \textsuperscript{\rm 1, \rm 2},\\{\bf Ying Tiffany He} \textsuperscript{\rm 1}{\bf, F. Richard Yu} \textsuperscript{\rm 3} \\
  \textsuperscript{\rm 1}College of Computer Science and Software Engineering, Shenzhen University, China \\
  \textsuperscript{\rm 2}Guangdong Laboratory of Artificial Intelligence and Digital Economy (SZ), China \\
  \textsuperscript{\rm 3}School of Information Technology, Carleton University, Canada \\
  }

\begin{document}
\maketitle
\begin{abstract}
Reinforcement learning with verifiable rewards improves reasoning in large language models (LLMs), but many methods still rely on large human-labeled datasets. While self-play reduces this dependency, it often lacks explicit planning and strong quality control, limiting stability in long-horizon multi-step reasoning. We present \textbf{SAGE} (\textbf{S}elf-evolving \textbf{A}gents for \textbf{G}eneralized reasoning \textbf{E}volution), a closed-loop framework where four agents: \emph{Challenger, Planner, Solver, and Critic}, co-evolve from a shared LLM backbone using only a small seed set. The Challenger continuously generates increasingly difficult tasks; the Planner converts each task into a structured multi-step plan; and the Solver follows the plan to produce an answer, whose correctness is determined by external verifiers. The Critic scores and filters both generated questions and plans to prevent curriculum drift and maintain training signal quality, enabling stable self-training. Across mathematics and code-generation benchmarks, SAGE delivers consistent gains across model scales, improving the Qwen-2.5-7B model by 8.9\% on LiveCodeBench and 10.7\% on OlympiadBench.
\end{abstract}

\section{Introduction}

Large language models (LLMs) have achieved remarkable advancements in reasoning tasks such as mathematics and coding through reinforcement learning (RL) techniques~\citep{guo2025deepseek_r1,sheng2025hybridflowflexibleefficient, sun2024llm}. However, these methods often depend on large-scale human-curated datasets for verifiable rewards, posing scalability challenges and limiting autonomous adaptation as models approach superhuman capabilities~\citep{absolute_zero_2025,chen_multi-agent_2025}.

\begin{figure}[h]
    \centering
    \includegraphics[width=\linewidth]{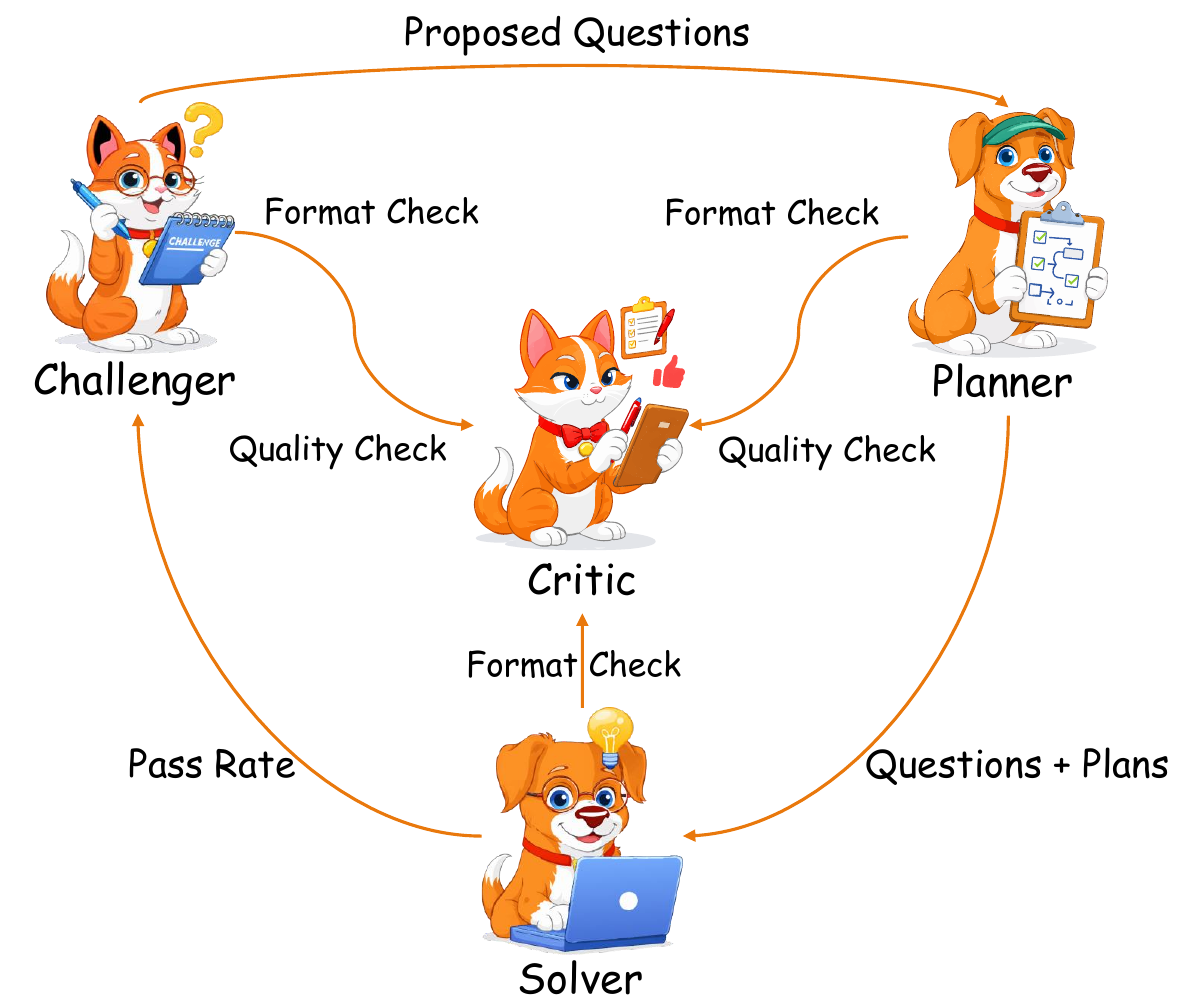}
    \caption{\textbf{Overview of the SAGE framework.} Four specialized agents—Challenger, Planner, Solver, and Critic—interact through quality filtering and format validation to enable closed-loop self-evolution.}
    \label{fig:overview_workflow}
\end{figure}

Recent efforts have explored self-play and multi-agent frameworks to enable self-evolution without extensive external data. For instance, self-play paradigms like SPIRAL~\citep{liu2025spiral} and Absolute Zero~\citep{absolute_zero_2025} leverage verifiable environments for autonomous improvement, while multi-agent systems such as MARS~\citep{yuan2025mars} and MAE~\citep{chen_multi-agent_2025} facilitate collaborative reasoning through role specialization. Despite these advances, existing approaches struggle with open-ended domains lacking robust verification and often fail to integrate planning for complex, multi-step tasks~\citep{huang2025rzeroselfevolvingreasoning,gao_survey_2025,yue2025doesreinforcementlearningreally}.

To address these gaps, we propose \textbf{SAGE} (\textbf{S}elf-evolving \textbf{A}gents for \textbf{G}eneralized reasoning \textbf{E}volution), a closed-loop multi-agent framework that enables LLMs to co-evolve in verifiable domains like math and coding using only minimal seed examples. As illustrated in Figure~\ref{fig:overview_workflow}, SAGE instantiates four specialized agents: a Challenger for task generation, a Planner for strategy outlining, a Solver for solution execution, and a Critic for quality assessment and format calibration. These agents interact adversarially, with the Challenger rewarded for difficulty and the Solver optimized via verifier-based correctness, forming a self-rewarding cycle trained end-to-end using task-relative policy gradients.

Through experiments on mathematics and coding benchmarks, SAGE demonstrates significant performance gains, outperforming baselines trained on human-curated datasets in sample efficiency and generalization.  We outline our contribution as follows: 
\begin{itemize}
    \item We design a scalable multi-agent framework for self-evolving LLMs in reasoning tasks.
    \item We propose a dual-role Critic mechanism ensuring task quality and solution verification.
    \item We conduct empirical evidence of effective co-evolution in math and code domains under few-example settings.
\end{itemize}

\begin{figure*}[!h]
    \centering
    \includegraphics[width=\textwidth]{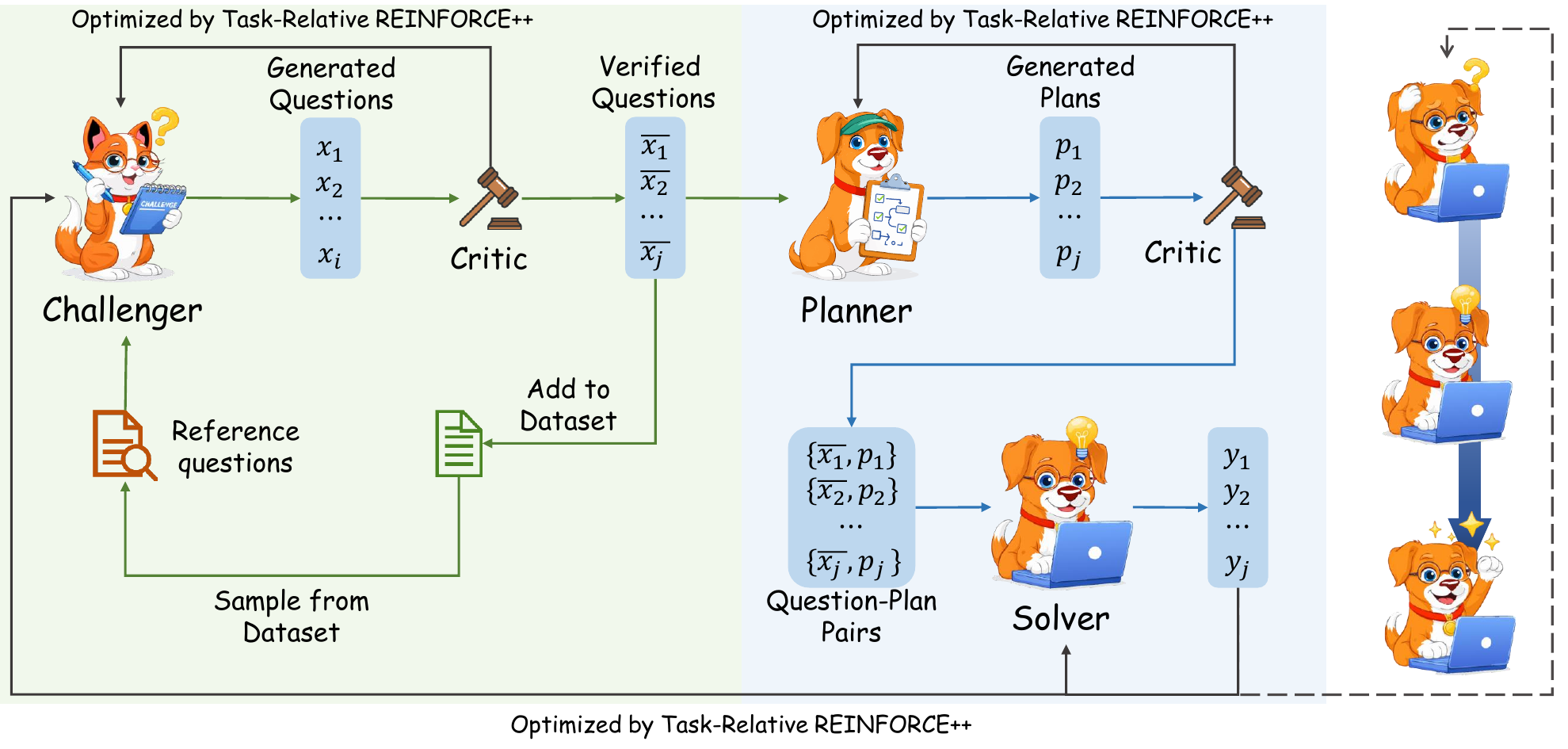}
    \caption{\textbf{The SAGE training pipeline.} (1) The Challenger generates questions from reference examples, filtered by the Critic for quality; (2) verified questions expand the dataset; (3) sampled questions are processed by the Planner and Solver to produce solutions; (4) all agents are jointly updated using Task-Relative REINFORCE++ with per-role advantage normalization.}

    \label{fig:training_loop}
\end{figure*}

\section{Related Work}
\noindent \textbf{Reinforcement Learning for LLM Reasoning.} Early work applied RL (e.g., PPO \citep{schulman2017proximal}) to language tasks, but recent research focuses on reinforcement learning with verifiable rewards (RLVR) for reasoning \citep{wan2025rema}. For example, DeepSeek-R1 \citep{guo2025deepseek_r1} shows that RLVR can extend an LLM’s reasoning capabilities on math by training from correctness signals. WebAgent-R1 \citep{wei2025webagent} is an end-to-end multi-turn RL framework that significantly boosts web navigation success using binary success rewards. Critic-free RL variants (e.g., GRPO \citep{guo2025deepseek_r1}) reduce training overhead, but typically still rely on human-curated or grounded environments. Recent work has systematically characterized agentic RL for LLMs, emphasizing capabilities like planning and self-improvement \citep{zhang_landscape_2025,wen2025reinforcement, evolver_self-evolving_2025}. In contrast, SAGE learns from self-generated, verifiable tasks with little external data.

\noindent \textbf{Multi-Agent LLM Systems.} LLM-based multi-agent frameworks facilitate complex tasks via role specialization. MetaGPT \citep{hong2024metagpt} encodes human-like workflows into a multi-agent assembly line, breaking down large tasks into subtasks among collaborating agents. CAMEL \citep{li_camel_2023} uses inception prompting to guide a society of role-playing agents, enabling study of cooperative behaviors in instruction-following tasks. MARS \citep{yuan2025mars} introduces a reinforcement learning framework where multi-agent self-play enhances strategic reasoning capabilities across cooperative and competitive tasks. These systems demonstrate that coordinating multiple LLM agents can enhance performance on complex tasks \citep{zhao2025stronger, zhu2025lamarl}. MARFT \citep{liao_marft_2025} applies multi-agent reinforcement fine-tuning to optimize LLM-based systems, and MALT \citep{motwani_malt_2025}, which divides reasoning into generation, verification, and refinement steps using heterogeneous agents. SAGE extends this line by instantiating distinct agents (Challenger, Planner, Solver, Critic) within one LLM and jointly training them with shared feedback.

\noindent \textbf{Self-Play and Self-Evolving Agents.} Recent works explore self-play and self-evolution to improve LLMs autonomously. The SPIRAL \citep{liu2025spiral} framework shows that self-play on zero-sum games can automatically induce generalizable reasoning strategies without human data. Absolute Zero \citep{absolute_zero_2025} generates its own coding problems and uses a code executor as a verifier to self-critique and solve them, achieving strong math and coding reasoning without external data. Agentic Self-Learning \citep{sun2025towardsagenticselflearningllms} is a closed-loop framework unifying task generation, policy execution, and reward modelling for LLM agents in search environments. Additional approaches include AgentEvolver \citep{zhai_agentevolver_2025} enables efficient self-evolving through curiosity-driven task generation and experience reuse, and Agent0 \citep{xia_agent0_2025}, which unleashes self-evolving agents via tool-integrated reasoning in a co-evolutionary curriculum-executor loop. While prior work has explored various components of self-evolving agents such as planning and task generation  \citep{gao_survey_2025, fang_comprehensive_2025, belle2025agents}, SAGE is distinguished by integrating planning and critic roles to decompose reasoning and jointly train all agents for improved stability and depth in math and code domains.

\section{Preliminaries}
\label{sec:preliminaries}
\noindent \textbf{Multi-Agent Reasoning in Verifiable Domains.} Let $\mathcal{M}_{\theta}$ denote an LLM parameterized by $\theta$. In \emph{role-based multi-agent reasoning}, multiple agents share a backbone model. Still, they are conditioned on different role instructions (e.g., proposer, planner, solver, evaluator) to enhance robustness via collaboration and decomposition~\cite{du2023improving, liang2023encouraging}. For a question $q$, agents produce structured answers $a$. In verifiable domains (mathematics, programming), a domain-specific verifier $V_{\mathrm{gt}}(q, a, v) \in [0,1]$ evaluates answer correctness given a reference $v$ (ground-truth or unit tests), enabling automatic reward computation without human annotation.

\noindent \textbf{Policy Gradient Optimization.} To enable self-evolution, we frame agent optimization as reinforcement learning, maximizing $J(\theta) = \mathbb{E}_{q \sim \mathcal{D}, o \sim \pi_\theta}[R(q,o)]$ where $\mathcal{D}$ is the task distribution, $R$ is the reward signal, and $o$ is the output. REINFORCE++~\citep{hu2025reinforcepp} is a critic-free method that computes the advantage as $A_{q,o}^t = r(q,o) - \beta_{\mathrm{kl}} \sum_{i=t}^{T} \mathrm{KL}(\pi_\theta \| \pi_{\mathrm{ref}})_i$ with KL penalty to a reference policy, and applies global-batch normalization: $A^{\mathrm{norm}} = (A - \mu_{\mathcal{B}})/(\sigma_{\mathcal{B}} + \epsilon)$. This stabilizes training and improves robustness across prompt distributions. To coordinate multiple agents with heterogeneous objectives, we adopt Task-Relative REINFORCE++~\citep{huang2025rzeroselfevolvingreasoning}, which applies per-role advantage normalization:
\begin{equation}
    A_{\mathrm{norm}}^{\mathrm{role}} = \frac{r - \mu_{\mathrm{role}}}{\sigma_{\mathrm{role}}+\epsilon},
\end{equation}
where $\mu_{\mathrm{role}}$ and $\sigma_{\mathrm{role}}$ are the mean and standard deviation computed over the corresponding role-specific batch.

\section{The SAGE Framework}

SAGE is a fully automated, self-iterative evolution framework requiring only a small seed set with automatic verification signals. SAGE instantiates four agents from a shared LLM backbone $\mathcal{M}_\theta$: (1) \textbf{Challenger} generates challenging tasks with verifiers; (2) \textbf{Planner} produces solution plans; (3) \textbf{Solver} outputs final answers; and (4) \textbf{Critic} evaluates quality and format compliance. These agents engage in continuous co-evolution, with the training workflow illustrated in Figure~\ref{fig:training_loop}.

In verifiable domains such as mathematics and programming, SAGE forms a closed-loop pipeline (challenge--plan--solve--criticize) that combines multi-agent interactions with verifier-based reward signals. The Challenger and Solver co-evolve adversarially: the Solver is rewarded for verified correctness, while the Challenger receives difficulty rewards when the Solver fails under verification, pushing the curriculum toward harder yet still solvable tasks. Quality filtering and verifier validation are applied to prevent dataset degradation and improve training stability.

\subsection{Reward Design and Normalization}
\noindent \textbf{Format reward.}
Across phases, SAGE applies a format reward $r_f\in[0,1]$ to stabilize self-training by enforcing required tags (e.g., \texttt{<question>}, \texttt{<answer>}, \texttt{<type>}, \texttt{<score>}). In practice, $r_f$ is a soft score (not strictly binary): missing tags yield low reward, redundant tags may receive partial credit, and empty outputs fall back to a neutral value (e.g., $0.5$).

\noindent \textbf{Score normalization.}
The Critic outputs scalar scores typically on a 1--10 scale inside \texttt{<score></score>}, which are normalized to $[0,1]$ by
\begin{equation}
\label{eq:norm}
\mathrm{Norm}(s)=
\begin{cases}
s, & 0\le s\le 1,\\
\frac{s-1}{9}, & 1 < s\le 10,\\
0.5, & \text{otherwise}.
\end{cases}
\end{equation}

\subsection{Challenger Agent Training}

The Challenger proposes verifiable tasks to drive the Solver's learning. During training, the Challenger policy $\pi_c$ is prompted with reference problems sampled from a small human-curated seed set $\mathcal{D}$ (about 500 examples across datasets), where each seed item includes a problem statement and its verifier (ground-truth answer or executable tests). Given a reference item $(q_{\mathrm{ref}}, v_{\mathrm{ref}})$, the Challenger generates a new problem $q$ and an associated verifier $v$ in a constrained format:
\begin{equation}
    (q,v) \sim \pi_c(\cdot \mid q_{\mathrm{ref}}, v_{\mathrm{ref}}; \theta),
\end{equation}
where $\theta$ represents the shared LLM parameters.

\noindent \textbf{Composite reward.}
The Challenger receives (i) a quality score $s_q\in[0,1]$ from the Critic (clarity, relevance, well-formedness), (ii) a difficulty reward computed from the Solver's verified success rate, and (iii) a format reward. Concretely, we estimate the Solver success by sampling $N_s$ answers and verifying them with $V_{\mathrm{gt}}$:
\begin{equation}
\begin{split}
 a_j \sim \pi_s(\cdot \mid q; \theta), \quad j=1,\ldots,N_s,\\
 \bar{s}_{\mathrm{gt}}(q,v)=\frac{1}{N_s}\sum_{j=1}^{N_s} V_{\mathrm{gt}}(q,a_j,v),\\
 r_d(q,v)=1-\bar{s}_{\mathrm{gt}}(q,v).
\end{split}
\end{equation}
Here, $V_{\mathrm{gt}}(q,a,v)\in[0,1]$ denotes the domain-specific verifier (e.g., exact-match/symbolic grading for math or test pass rate for code), $\pi_s$ denotes the Solver policy (formally introduced in Section \ref{subsec:sovler}).

The Challenger reward is computed as
\begin{equation}
\label{eq:challenger_reward}
 r_c(q,v)=\tfrac{1}{3}s_q(q)+\tfrac{1}{3}r_d(q,v)+\tfrac{1}{3}r_f(o_c),
\end{equation}
where $o_c$ (resp. $o_p$, $o_s$, $o_{cr}$) denotes the raw textual output of the Challenger (resp. Planner, Solver, Critic).

\noindent \textbf{Quality filtering and difficulty suppression.}
To prevent dataset degradation, we filter low-quality questions with a threshold $\alpha$ (in this paper, $\alpha=0.7$), and also validate the generated verifier (e.g., parsable and executable for code tests). Only candidates that satisfy both criteria are added to $\mathcal{D}$. Moreover, for $s_q<\alpha$, we suppress the difficulty term to avoid rewarding ``hard'' but ill-posed tasks and use
\begin{equation}
 r_c(q,v)=\tfrac{1}{2}s_q(q)+\tfrac{1}{2}r_f(o_c).
\end{equation}
This stabilizes long-horizon self-training and mitigates reward collapse.

\begin{algorithm}[t]
\footnotesize
\caption{Training Process of SAGE}
\label{algo:SAGE}
\begin{algorithmic}[1]
\Require Base LLM $\pi_{\text{base}}$, iterations $T$, thresholds $\alpha,\beta$,   

sample size $N_s$

\State Init agents $\pi_c,\pi_p,\pi_s,\pi_{cr}$ from $\pi_{\text{base}}$
\State Init dataset $\mathcal{D}\leftarrow\mathcal{D}_0$ (each item has verifier)

\For{$t=1$ to $T$}
\State Sample $(q_{\rm ref},v_{\rm ref})\sim\mathcal{D}$  \Comment{\textbf{(1) Challenge Phase}}
\State $(q_t,v_t)\leftarrow\pi_c(\cdot\mid q_{\rm ref},v_{\rm ref})$
\State $s_q\leftarrow \mathrm{Norm}(\pi_{cr}(q_t))$; validate $v_t$
\State Sample $a_j\sim\pi_s(\cdot\mid q_t)$ for $j=1,\ldots,N_s$
\State $\bar{s}_{\mathrm{gt}}\leftarrow \frac{1}{N_s}\sum_{j=1}^{N_s} V_{\mathrm{gt}}(q_t,a_j,v_t)$; $r_d\leftarrow 1-\bar{s}_{\mathrm{gt}}$
\If{$s_q\ge\alpha$ \textbf{and} $v_t$ valid}
  \State $\mathcal{D}\leftarrow\mathcal{D}\cup\{(q_t,v_t)\}$; $r_c\leftarrow \tfrac{1}{3}s_q+\tfrac{1}{3}r_d+\tfrac{1}{3}r_f(o_c)$
\Else
  \State $r_c\leftarrow \tfrac{1}{2}s_q+\tfrac{1}{2}r_f(o_c)$ 
\EndIf

\State \Comment{\textbf{(2) Plan--Solve Phase}}
\State Sample $(q,v)\sim\mathcal{D}$; $p_t\leftarrow\pi_p(\cdot\mid q)$
\State $s_p\leftarrow \mathrm{Norm}(\pi_{cr}(q,p_t))$
\If{$s_p\ge\beta$}
  \State $a_t\leftarrow\pi_s(\cdot\mid q,p_t;\theta)$; $\tilde{s}_p\leftarrow s_p$
\Else
  \State $a_t\leftarrow\pi_s(\cdot\mid q,\emptyset;\theta)$; $\tilde{s}_p\leftarrow 0$
\EndIf
\State $s_{\mathrm{gt}}\leftarrow V_{\mathrm{gt}}(q,a_t,v)$
\State $r_p\leftarrow \lambda_{\rm plan}s_p+\lambda_f r_f(o_p)$; 
\State $r_s\leftarrow w_p \tilde{s}_p+w_c s_{\mathrm{gt}}+w_f r_f(o_s)$
\State $r_{cr}\leftarrow r_f(o_{cr})$ \Comment{\textbf{(3) Joint Update}}
\State Update $\pi_c,\pi_p,\pi_s,\pi_{cr}$ using $r_c,r_p,r_s,r_{cr}$

\EndFor
\end{algorithmic}
\end{algorithm}

\subsection{Planner Agent Training}

The Planner $\pi_p$ generates a structured plan $p$ for a given question $q$, encapsulated in \texttt{<plan></plan>} tags. The Critic evaluates the plan quality to produce a normalized score $s_p\in[0,1]$.
\begin{equation}
 p \sim \pi_p(\cdot \mid q;\theta), \quad s_p=\mathrm{Norm}\big(\mathrm{Critic}(q,p)\big).
\end{equation}
If $s_p$ meets a gating threshold (in this paper, $\beta=0.3$), the plan is provided to the Solver; otherwise, the Solver answers directly.

For optimizing the Planner, we use a composite reward that combines plan quality and format compliance:
\begin{equation}
 r_p = \lambda_{\mathrm{plan}}\, s_p + \lambda_f\, r_f(o_p),
\end{equation}
where $\lambda_{\mathrm{plan}}$ and $\lambda_f$ are weighting coefficients (we use $\lambda_{\mathrm{plan}}=\lambda_f=0.5$ by default).

\subsection{Solver Agent Training}
\label{subsec:sovler}
The Solver agent is tasked with generating final answers based on the given question $q$ and the plan $p$ (if the plan passes Critic gating). The Solver policy $\pi_s$ produces an answer $a$, typically wrapped in \texttt{<answer></answer>} tags or Markdown blocks:
\begin{equation}
    a \sim \pi_s(\cdot \mid q, \tilde{p}; \theta),\quad
    \tilde{p}=
    \begin{cases}
    p,& s_p\ge\beta,\\
    \emptyset,& s_p<\beta.
    \end{cases}
\end{equation}

\noindent \textbf{Verifier-based composite reward (plan, correctness, format).}
Solver correctness is computed by automatic verification in the target domain (symbolic/metric-based grading for math, or execution-/test-based validation for code), yielding $s_{\mathrm{gt}}\in[0,1]$.
We combine plan quality, verified correctness, and format adherence as
\begin{equation}
\begin{split}
\label{eq:solver_reward}
\tilde{s}_p =
\begin{cases}
 s_p, & s_p\ge\beta,\\
 0, & s_p<\beta,
\end{cases}
\end{split}
\end{equation}
\begin{equation}
    \begin{split}
         r_s = w_p\, \tilde{s}_p + w_c\, s_{\mathrm{gt}} + w_f\, r_f(o_s),\\
 w_p+w_c+w_f=1.
    \end{split}
\end{equation}
In this paper, we use $(w_p,w_c,w_f)=(0.2,0.6,0.2)$ as the default setting.
If the plan score is unavailable (e.g., when the planning module is disabled), we fall back to a simpler mixture of verified correctness and format (e.g., $\frac{1}{2}s_{\mathrm{gt}}+\frac{1}{2}r_f$) to maintain robustness.

In adversarial interaction with the Challenger, Solver failures under ground-truth verification contribute to the Challenger's difficulty reward (Eq.~\ref{eq:challenger_reward}), forming a co-evolutionary loop that progressively pushes the curriculum toward harder yet solvable problems.

\subsection{Critic: Scoring and Format Calibration}

The Critic provides two types of signals: (1) soft format rewards $r_f\in[0,1]$ by checking required tags, and (2) quality scores for Challenger questions ($s_q$) and Planner plans ($s_p$), normalized via Eq.~\ref{eq:norm}. Importantly, in the verifiable setting, correctness is determined by the external verifier $V_{\mathrm{gt}}$ rather than the Critic.

The Critic policy $\pi_{cr}$ outputs a scalar score deterministically:
\begin{equation}
    s \sim \pi_{cr}(\cdot \mid x; \theta),
\end{equation}
where $x \in \{(q, \cdot), (q, p)\}$ denotes the evaluation context (either a question alone or a question-plan pair). Optionally, we calibrate the Critic with a lightweight format-consistency objective
\begin{equation}
r_{cr}=r_f(o_{cr}),
\end{equation}
which reduces parsing failures and improves stability of downstream reward computation.

\begin{table*}[t]
\centering
\small
\setlength{\tabcolsep}{2pt}
\begin{tabular}{l|ccc|cccccc|ccc}
\toprule
\textbf{Method} &
\textbf{HEval+} &\textbf{ MBPP+} &\textbf{ LCB$^{v1-5}$} &
\textbf{GSM8K} & \textbf{Math} & \textbf{AI24} & \textbf{AI25} & \textbf{AMC} & \textbf{Olympiad} &
\textbf{C Avg.} &\textbf{ M Avg.} & \textbf{O Avg.} \\
\midrule
\multicolumn{13}{c}{\textbf{Qwen‑2.5‑3B-Instruct}} \\
\midrule
Base Model & 68.3 & 60.6 & 12.0 & 84.6 & 60.4 & 3.3 & \textbf{6.7} & \textbf{40.0} & 28.0 & 46.9 & 37.2 & 40.4 \\
AZR        & \textbf{68.9} & 61.4 & 15.0 & 81.2 & 62.4 & 3.3 & 3.3 & 35.0 & 28.9 & 48.4 & 35.7 & 39.9 \\
MAE        & 68.3 & 61.1 & 15.9 & 82.2 & 65.8 & 3.3 & 3.3 & 32.5 & \textbf{32.5} & 48.4 & 36.6 & 40.5 \\
\texttt{SAGE} & \textbf{68.9} & \textbf{62.4} & \textbf{16.9} & \textbf{85.5} & \textbf{66.2} & \textbf{6.7} & \textbf{6.7} & 35.0 & 29.8 & \textbf{49.4} & \textbf{38.3} & \textbf{42.0} \\
\midrule
\multicolumn{13}{c}{\textbf{Qwen-2.5-7B-Instruct}} \\
\midrule
Base Model & 73.2 & 65.3 & 17.5 & 91.7 & 75.1 & \textbf{13.3} & 6.7 & \textbf{57.5} & 28.0 & 52.0 & 45.4 & 47.6 \\
AZR        & 71.3 & \textbf{69.1} & 25.3 & \textbf{92.8} & \textbf{76.2} & 10.0 & \textbf{13.3} & 50.0 & 38.5 & 55.2 & 46.8 & 49.6 \\
MAE        & \textbf{76.2} & 65.3 & 23.3 & 91.7 & \textbf{76.2} & \textbf{13.3} & \textbf{13.3} & 42.5 & 32.7 & 54.9 & 45.0 & 48.3 \\
\texttt{SAGE}      & \textbf{76.2} & 64.0 & \textbf{26.4} & 92.2 & 74.7 & \textbf{13.3} & \textbf{13.3} & 52.5 & \textbf{38.7} & \textbf{55.5} & \textbf{47.5} & \textbf{50.1} \\
\midrule
\multicolumn{13}{c}{\textbf{Qwen-3-4B-Base}} \\
\midrule
Base Model & \textbf{76.8} &\textbf{ 65.3} & 21.5 & \textbf{94.5} & 87.0 & \textbf{16.7} & \textbf{13.3} & \textbf{77.5} & \textbf{49.0} & 54.5 & \textbf{56.3} & 55.7 \\
AZR   & 74.4 & 65.0 & 26.1 & 89.3 & 76.2 & 10.0 & \textbf{13.3} & 50.0 & 41.5 & 55.2 & 46.7 & 49.5 \\
MAE        & 76.2 & \textbf{65.3} & 24.2 & \textbf{94.5} & \textbf{92.0} & 13.3 & 10.0 & 70.0 & 43.7 & 55.2 & 53.9 & 54.4 \\
\texttt{SAGE}   & 75.6 & 62.4 & \textbf{30.6 }& 94.3 & 91.0 & \textbf{16.7} & 10.0 & 75.0 & 47.9 & \textbf{56.2} & 55.8 & \textbf{55.9} \\
\bottomrule
\end{tabular}
\caption{\textbf{Main results on reasoning benchmarks.} Comparison of post-training methods across three model scales. We report pass@1 accuracy (\%) on code generation (HumanEval+, MBPP+, LiveCodeBench) and mathematical reasoning (GSM8K, MATH, AIME 2024, AIME 2025, AMC, and OlympiadBench). C Avg., M Avg., and O Avg. denote the mean scores over code, math, and all benchmarks. SAGE achieves the best overall performance across all three model backbones. Bold indicates best per LLM backbone.}

\label{tab:qwen25_results}
\end{table*}

\subsection{Multi-Agent Co-Training}

A training step in SAGE comprises: (1) \textbf{Challenger Phase} to generate verifiable candidate tasks and expand $\mathcal{D}$ with quality-and-verifier filtering; (2) \textbf{Plan--Solve Phase} where the Planner generates a single plan scored by the Critic and the Solver is optimized using the verifier-based reward in Eq.~\ref{eq:solver_reward}; (3) \textbf{Critic Phase} (optional) for format calibration; and (4) \textbf{Synchronized Update} that jointly updates the shared backbone using Task-Relative REINFORCE++ with per-role advantage normalization (see Section \ref{sec:preliminaries}).

\section{Experiments}

\subsection{Experimental Setup}
\textbf{Training details}
Our framework is implemented based on VeRL\citep{sheng2025hybridflowflexibleefficient}, and we evaluate it using the Qwen2.5-3B-Instruct, Qwen2.5-7B-Instruct, and Qwen3-4B-Base models\citep{yang2024qwen2,yang2025qwen3technicalreport}. All agents are initialized from their corresponding base models. We apply LoRA \citep{hu2022lora} with rank 128 and a learning rate of 3e-6. Additional hyperparameter settings are provided in Table \ref{tab:hyperparams}.

\noindent \textbf{Baseline Methods.} To comprehensively assess the effectiveness of the proposed SAGE framework, we conduct experiments on several representative foundation models, including Qwen2.5-3B-Instruct, Qwen2.5-7B-Instruct, and Qwen3-4B-Base. For each model, we report results for both the original checkpoint and the corresponding variant fine-tuned with SAGE. In addition, we include Absolute-Zero-Reasoning (AZR) \citep{absolute_zero_2025} and Multi-Agent Evolve (MAE) \citep{chen_multi-agent_2025} as alternative training baselines. Specifically, each model is trained for 200 steps under AZR. For MAE, we adopt the half-reference setting and train each model for 200 steps.

\noindent \textbf{Training and Evaluation Datasets.} Our training set comprises 500 instances sampled from MATH \citep{hendrycks2020measuring}, GSM8K \citep{cobbe2021training}, HumanEval \citep{chen2021evaluatinglargelanguagemodels}, and MBPP \citep{austin2021program}, with detailed statistics in Appendix~\ref{app:dataset}. We evaluate on two domains: (1) \textit{Mathematical Reasoning}: GSM8K and MATH (in-distribution, ID), along with four competition-level benchmarks—AIME'24, AIME'25, OlympiadBench \citep{He2024OlympiadBenchAC}, and AMC'23 \citep{hendrycks2021measuringmathematicalproblemsolving}—as out-of-distribution (OOD) tests. (2) \textit{Code Generation}: HumanEval+ and MBPP+ evaluated via Evalplus \citep{evalplus} (ID), and LiveCodeBench \citep{jain2024livecodebench} v1–v5 (May 2023–February 2025) for OOD assessment. We report the accuracy (pass@1) based on greedy decoding across all benchmarks.

\subsection{Main Results}
Table~\ref{tab:qwen25_results} presents the performance of SAGE and baseline methods across code generation and mathematical reasoning benchmarks on three model backbones.

\noindent \textbf{Consistent Improvements Across Model Scales.} SAGE achieves the highest Overall Avg. on both Qwen-2.5-3B-Instruct (42.0\%) and Qwen-2.5-7B-Instruct (50.1\%), outperforming all baselines including AZR and MAE. On the 3B model, SAGE improves upon the base model by 1.6\% overall, with notable gains on in-distribution benchmarks (GSM8K: 84.6\% $\rightarrow$ 85.5\%; MATH: 60.4\% $\rightarrow$ 66.2\%). Similarly, on the 7B model, SAGE yields a 2.5\% improvement over the base model in Overall Avg., demonstrating consistent effectiveness across model scales.

\begin{table}[t]
\centering
\small

\begin{tabular}{llcc}
\toprule
\textbf{Backbone} & \textbf{Method} & \textbf{ID Avg.} & \textbf{OOD Avg.} \\
\midrule
\multirow{4}{*}{Qwen-2.5-3B} & Base Model & 68.4 & 18.0 \\
& AZR & 68.5 & 17.1 \\
& MAE & 69.4 & 17.5 \\
& \texttt{SAGE} & \textbf{70.8} & \textbf{19.0} \\
\midrule
\multirow{4}{*}{Qwen-2.5-7B} & Base Model & 76.3 & 24.6 \\
& AZR & \textbf{77.4} & 27.4 \\
& MAE & 76.8 & 25.0 \\
& \texttt{SAGE} & \textbf{77.4} & \textbf{28.8} \\
\midrule
\multirow{4}{*}{Qwen-3-4B} & Base Model & 80.9 & 35.6 \\
& AZR & 76.2 & 28.2 \\
& MAE & \textbf{82.0} & 32.2 \\
& \texttt{SAGE} & 80.8 & \textbf{36.0} \\
\bottomrule
\end{tabular}

\caption{\textbf{ID and OOD generalization comparison.} SAGE consistently improves OOD performance (+4.2\% on 7B) without sacrificing in-distribution accuracy.}

\label{tab:id_ood}
\end{table}

\begin{table*}[h]
\centering
\small
\setlength{\tabcolsep}{1.6pt}
\begin{tabular}{l|ccc|cccccc|ccc}
\toprule
\textbf{Method} &
\textbf{HEval+} &\textbf{ MBPP+} &\textbf{ LCB$^{v1-5}$ }&
\textbf{GSM8K} & \textbf{Math} & \textbf{AI24} & \textbf{AI25} & \textbf{AMC} & \textbf{Olympiad} &
\textbf{C Avg.} & \textbf{M Avg.} & \textbf{O Avg.} \\
\midrule

SAGE$^\text{(full implementation)}$    & \textbf{68.9} & 62.4 & \textbf{16.9} & 85.5 & \textbf{66.2} & \textbf{6.7} & \textbf{6.7} & 35.0 & \textbf{29.8} & \textbf{49.4} & \textbf{38.3} & \textbf{42.0} \\
SAGE$^\text{(w/o challenger training)}$       & 66.5 & 61.3 & 9.0 & \textbf{86.7} & 65.5 & 0.0 & 3.3 & 35.0 & 28.0 & 45.6 & 36.4 & 39.5 \\
SAGE$^\text{(w/o solver training)}$           & 67.7  & \textbf{64.3}  & 9.0 & 81.2 & 60.4 & 3.3 & 0.0 & 30.0 & 28.0 & 47.0 & 33.8 & 38.2 \\
SAGE$^\text{(w/o critic training)}$           & 66.5 & 53.7 & 14.1 & 86.0 & 65.9 & 3.3 & \textbf{6.7} & \textbf{40.0} & 27.4 & 44.8 & 38.2 & 40.4 \\
\midrule
\end{tabular}
\caption{\textbf{Ablation study of SAGE components on Qwen-2.5-3B.} We evaluate the impact of removing individual agent training while keeping other components active.}
\label{tab:ablation_results}
\end{table*}

\noindent \textbf{Strong Out-of-Distribution Generalization.} A key strength of SAGE lies in its generalization to out-of-distribution benchmarks. As shown in Table~\ref{tab:id_ood}, SAGE achieves the best or near-best OOD Avg. across all three backbones (19.0\%, 28.8\%, and 36.0\% respectively), while maintaining competitive ID Avg. scores. This balanced improvement is particularly evident on Qwen-2.5-7B, where SAGE improves OOD Avg. by 4.2\% over the base model while preserving strong in-distribution performance. On LiveCodeBench specifically, SAGE achieves the best performance across all three backbones (16.9\%, 26.4\%, and 30.6\%), substantially outperforming both base models and other post-training methods. For mathematical reasoning, SAGE maintains competitive performance on competition-level benchmarks such as OlympiadBench, where it achieves 38.7\% (+10.7\% over base) on Qwen-2.5-7B.

\noindent \textbf{Comparison with Baselines.} While AZR and MAE show improvements on certain individual benchmarks, they exhibit inconsistent gains and occasional performance degradation. For instance, AZR on Qwen-3-4B-Base leads to a significant drop in Math Avg. (56.3\% $\rightarrow$ 46.7\%). In contrast, SAGE maintains more balanced improvements across both domains without sacrificing performance on any benchmark group.

\noindent \textbf{Results on Qwen-3-4B.} On this stronger backbone, the base model already achieves high performance (Overall Avg. 55.7\%). Nevertheless, SAGE attains the highest Code Avg. (56.2\%) and remains competitive overall (55.9\%), with particularly strong gains on LiveCodeBench (21.5\% $\rightarrow$ 30.6\%, +9.1\%). This suggests that SAGE continues to provide meaningful improvements even when applied to capable base models.

\subsection{Ablations Studies and Analyses}

\noindent \textbf{Ablation Study.} To understand the contribution of each agent, we conduct ablation experiments by selectively disabling the training of individual roles while keeping the remaining components active. As shown in Table~\ref{tab:ablation_results}, the full SAGE implementation achieves the highest overall average (42.0\%), and removing any single agent leads to performance degradation.

Disabling Challenger training results in a notable drop in code benchmarks, particularly on LiveCodeBench (16.9\% $\rightarrow$ 9.0\%), indicating that curriculum generation is essential for out-of-distribution generalization. Similarly, removing Solver training causes the largest overall decline (O Avg. 38.2\%), with substantial drops on both GSM8K (85.5\% $\rightarrow$ 81.2\%) and MATH (66.2\% $\rightarrow$ 60.4\%), confirming that the Solver is the primary driver of reasoning capability. Interestingly, excluding Critic training yields competitive math performance (M Avg. 38.2\%) but degrades code benchmarks (C Avg. 44.8\%), suggesting that the Critic's quality filtering is more critical for code generation where output format and correctness are tightly coupled.

These results validate that all three trainable agents contribute complementarily to SAGE's overall effectiveness, with the Challenger--Solver interaction forming the core co-evolutionary loop and the Critic providing essential quality control.

\begin{figure}[!h]
    \centering
    \includegraphics[width=\linewidth]{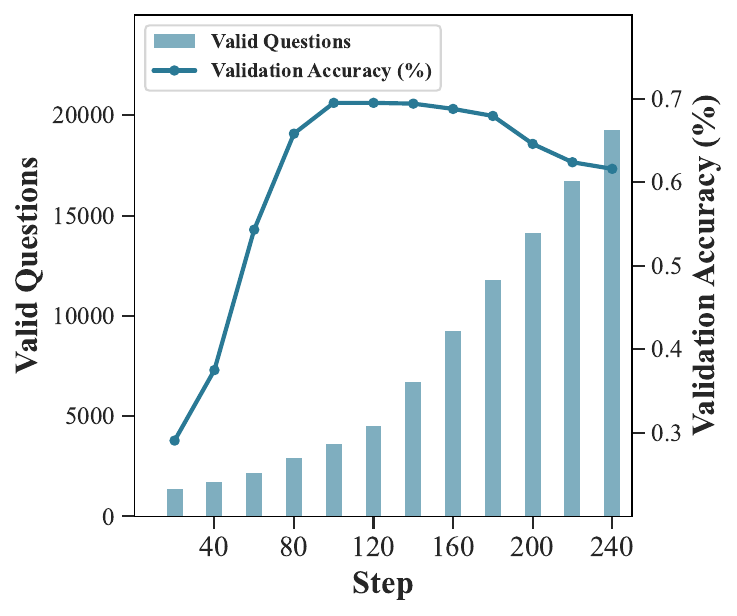}
    \caption{\textbf{Training dynamics on Qwen-2.5-3B.} The Challenger steadily expands the question pool (bars) throughout training, while validation accuracy (line) reaches peak performance around step 100--120 before gradual decline, suggesting potential over-specialization on the self-generated curriculum.}

    \label{fig:training_dynamics}
\end{figure}

\begin{figure*}[ht]
    \centering
    \includegraphics[width=\textwidth]{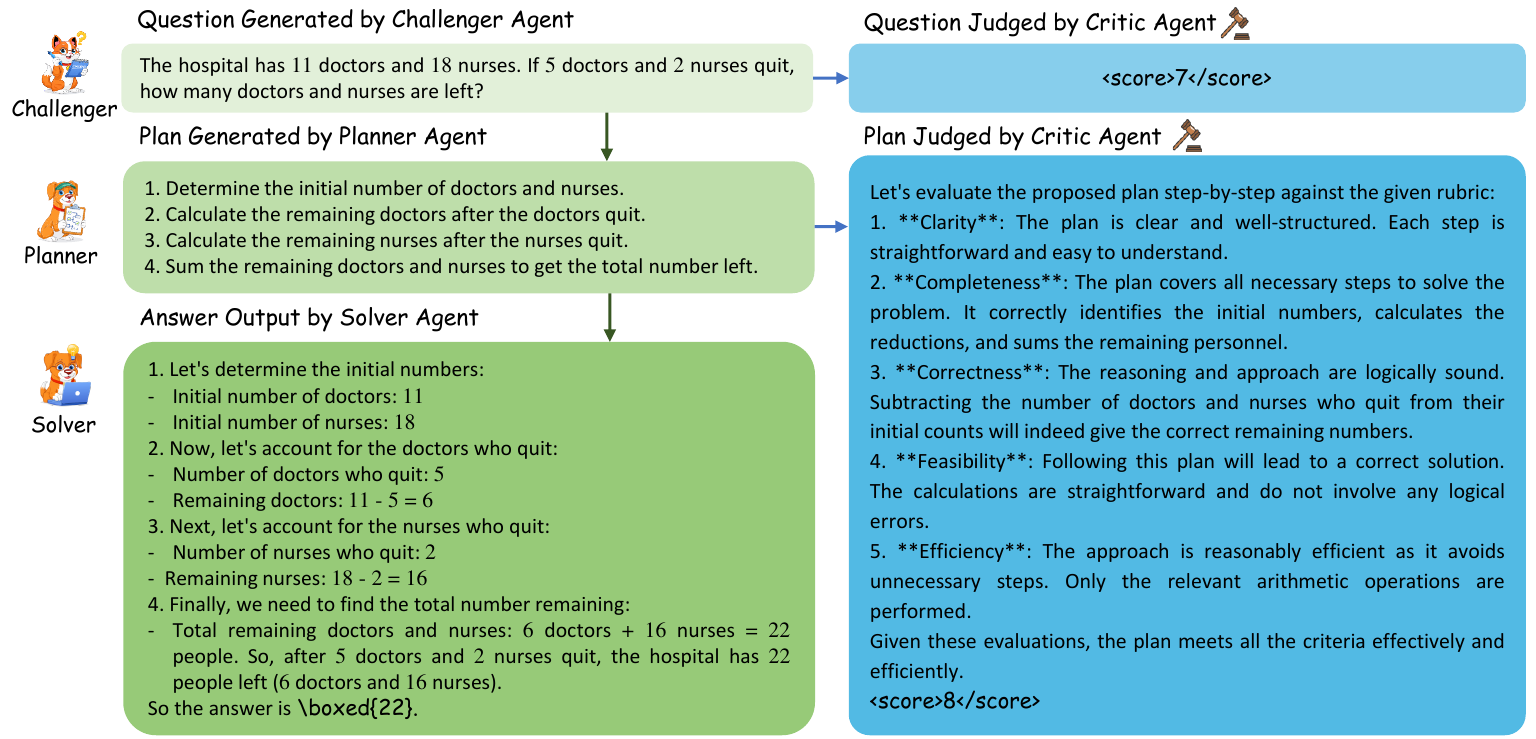}
    \caption{\textbf{Qualitative case study.} The Challenger generates a math word problem, the Planner decomposes it into structured steps, the Solver executes the plan to produce the final answer, and the Critic provides quality scores for both the question and the plan.}
    \label{fig:case_study}
\end{figure*}

\noindent \textbf{Training Dynamics Analysis.} To gain deeper insights into the self-evolution process, we analyze the training dynamics of SAGE on Qwen-2.5-3B-Instruct, as shown in Figure~\ref{fig:training_dynamics}.

The validation accuracy (line) exhibits a characteristic learning curve. During the initial phase (steps 0--80), the model demonstrates rapid improvement from 29.1\% to 65.8\%, reflecting efficient knowledge acquisition from the multi-agent co-evolutionary training. The accuracy reaches its peak of  69.5\% around step 100--140, representing the optimal balance between task difficulty and model capability. Beyond this point, we observe a gradual decline to 61.6\% by step 240, suggesting that prolonged training may lead to over-specialization on the self-generated curriculum. This motivates our choice of reporting results around step 100 in the main experiments.

Meanwhile, the cumulative number of valid questions (bars) grows steadily throughout training, expanding from 1,136 to 20,532 by step 250, an 18-fold increase from the seed set. Notably, the growth rate accelerates around step 120--130, coinciding with peak validation accuracy, suggesting that a well-trained Challenger produces questions that pass the quality threshold $\alpha=0.7$ at an increasing rate. The continued growth of the question pool despite declining accuracy after step 120 suggests that increased quantity alone does not ensure better performance, highlighting the importance of curriculum diversity and difficulty calibration. Nevertheless, this trend demonstrates SAGE’s ability to autonomously scale its training data without human intervention.

\noindent \textbf{Qualitative Analysis.} Figure~\ref{fig:case_study} illustrates the collaborative reasoning process of SAGE. The Challenger generates a well-formed arithmetic problem involving subtraction across two categories. The Planner decomposes this into four sequential steps, progressing from initial value identification to final summation. Guided by this structured plan, the Solver executes each step systematically and arrives at the correct answer. The Critic evaluates both outputs, assigning scores of 7 and 8 based on clarity, completeness, and logical soundness. This example highlights how role specialization enables effective division of labor: task generation, strategic planning, solution execution, and quality assessment operate as distinct yet coordinated functions within a unified training loop.

\section{Conclusion}

We introduce SAGE, a multi-agent self-evolution framework where four specialized agents: Challenger, Planner, Solver, and Critic, co-evolve through adversarial yet collaborative dynamics. Starting from minimal seed examples, SAGE autonomously expands its training curriculum while maintaining quality via critic-based filtering. Experiments demonstrate consistent improvements across model scales, with strong out-of-distribution generalization on competition-level benchmarks. These results highlight a scalable and effective pathway for evolving capable reasoning agents while reducing dependency on human-curated supervision.

\section{Limitations}

Among the limitations of our work, firstly, SAGE operates in verifiable domains where correctness can be automatically determined through ground-truth answers or executable tests. Extending the framework to open-ended tasks with subjective evaluation criteria, potentially through learned reward models, remains an interesting direction for future work. Secondly, although SAGE significantly reduces reliance on large-scale annotations, it still requires a small seed set (500 examples) to bootstrap the self-evolution process. Investigating strategies to further minimize seed requirements could broaden applicability to extremely low-resource scenarios. Thirdly, our evaluation focuses on mathematical reasoning and code generation benchmarks. Future exploration of other structured reasoning domains, such as logical reasoning or scientific problem solving, could offer valuable insights and validate the generalizability of our multi-agent architecture. Additionally, as with standard self-training approaches, monitoring training dynamics and applying early stopping is advisable to ensure optimal performance.

\bibliography{custom}

@misc{chen2021evaluatinglargelanguagemodels,
      author={Mark Chen and Jerry Tworek and Heewoo Jun and Qiming Yuan and Henrique Ponde de Oliveira Pinto and Jared Kaplan and Harri Edwards and Yuri Burda and Nicholas Joseph and Greg Brockman and Alex Ray and Raul Puri and Gretchen Krueger and Michael Petrov and Heidy Khlaaf and Girish Sastry and Pamela Mishkin and Brooke Chan and Scott Gray and Nick Ryder and Mikhail Pavlov and Alethea Power and Lukasz Kaiser and Mohammad Bavarian and Clemens Winter and Philippe Tillet and Felipe Petroski Such and Dave Cummings and Matthias Plappert and Fotios Chantzis and Elizabeth Barnes and Ariel Herbert-Voss and William Hebgen Guss and Alex Nichol and Alex Paino and Nikolas Tezak and Jie Tang and Igor Babuschkin and Suchir Balaji and Shantanu Jain and William Saunders and Christopher Hesse and Andrew N. Carr and Jan Leike and Josh Achiam and Vedant Misra and Evan Morikawa and Alec Radford and Matthew Knight and Miles Brundage and Mira Murati and Katie Mayer and Peter Welinder and Bob McGrew and Dario Amodei and Sam McCandlish and Ilya Sutskever and Wojciech Zaremba},
title = {{Evaluating Large Language Models Trained on Code}},
  year = {2021},
  eprint = {2107.03374},
  archivePrefix = {arXiv},
  primaryClass = {cs.LG},
  url = {https://arxiv.org/abs/2107.03374},
  doi = {10.48550/arXiv.2107.03374}
}

@misc{austin2021program,
author = {Austin, Jacob and Odena, Augustus and Nye, Maxwell and Bosma, Maarten and Michalewski, Henryk and Dohan, David and Jiang, Ellen and Cai, Carrie and Terry, Michael and Le, Quoc and Sutton, Charles},
  title = {{Program Synthesis with Large Language Models}},
  year = {2021},
  eprint = {2108.07732},
  archivePrefix = {arXiv},
  primaryClass = {cs.PL},
  url = {https://arxiv.org/abs/2108.07732},
  doi = {10.48550/arXiv.2108.07732}
}

@inproceedings{sheng2025hybridflowflexibleefficient,
  author = {Sheng, Guangming and Zhang, Chi and Ye, Zilingfeng and Wu, Xibin and Zhang, Wang and Zhang, Ru and Peng, Yanghua and Lin, Haibin and Wu, Chuan},
  title = {{HybridFlow: A Flexible and Efficient RLHF Framework}},
  booktitle = {Proceedings of the Twentieth European Conference on Computer Systems},
  year = {2025},
  address = {Rotterdam, The Netherlands},
  publisher = {ACM},
  pages = {1279--1297},
  doi = {10.1145/3689031.3696075},
  url = {https://doi.org/10.1145/3689031.3696075}
}

@misc{chen_multi-agent_2025,
author = {Chen, Yixing and Wang, Yiding and Zhu, Siqi and Yu, Haofei and Feng, Tao and Zhang, Muhan and Patwary, Mostofa and You, Jiaxuan},
  title = {{Multi-Agent Evolve: LLM Self-Improve through Co-evolution}},
  year = {2025},
  eprint = {2510.23595},
  archivePrefix = {arXiv},
  primaryClass = {cs.AI},
  url = {https://arxiv.org/abs/2510.23595},
  doi = {10.48550/arXiv.2510.23595}
}

@misc{cobbe2021training,
author = {Cobbe, Karl and Kosaraju, Vineet and Bavarian, Mohammad and Chen, Mark and Jun, Heewoo and Kaiser, Lukasz and Plappert, Matthias and Tworek, Jerry and Hilton, Jacob and Nakano, Reiichiro and Hesse, Christopher and Schulman, John},
  title = {{Training Verifiers to Solve Math Word Problems}},
  year = {2021},
  eprint = {2110.14168},
  archivePrefix = {arXiv},
  primaryClass = {cs.LG},
  url = {https://arxiv.org/abs/2110.14168},
  doi = {10.48550/arXiv.2110.14168}
}

@misc{du2023improving,
author = {Du, Yilun and Li, Shuang and Torralba, Antonio and Tenenbaum, Joshua B. and Mordatch, Igor},
  title = {{Improving Factuality and Reasoning in Language Models through Multiagent Debate}},
  year = {2023},
  eprint = {2305.14325},
  archivePrefix = {arXiv},
  primaryClass = {cs.CL},
  url = {https://arxiv.org/abs/2305.14325},
  doi = {10.48550/arXiv.2305.14325}
}

@misc{fang_comprehensive_2025,
author = {Fang, Jinyuan and Peng, Yanwen and Zhang, Xi and Wang, Yingxu and Yi, Xinhao and Zhang, Guibin and Xu, Yi and Wu, Bin and Liu, Siwei and Li, Zihao and Ren, Zhaochun and Aletras, Nikos and Wang, Xi and Zhou, Han and Meng, Zaiqiao},
  title = {{A Comprehensive Survey of Self-Evolving AI Agents: A New Paradigm Bridging Foundation Models and Lifelong Agentic Systems}},
  year = {2025},
  eprint = {2508.07407},
  archivePrefix = {arXiv},
  primaryClass = {cs.AI},
  url = {https://arxiv.org/abs/2508.07407},
  doi = {10.48550/arXiv.2508.07407}
}

@misc{gao_survey_2025,
author = {Gao, Huan-ang and Geng, Jiayi and Hua, Wenyue and Hu, Mengkang and Juan, Xinzhe and Liu, Hongzhang and Liu, Shilong and Qiu, Jiahao and Qi, Xuan and Wu, Yiran and Wang, Hongru and Xiao, Han and Zhou, Yuhang and Zhang, Shaokun and Zhang, Jiayi and Xiang, Jinyu and Fang, Yixiong and Zhao, Qiwen and Liu, Dongrui and Ren, Qihan and Qian, Cheng and Wang, Zhenhailong and Hu, Minda and Wang, Huazheng and Wu, Qingyun and Ji, Heng and Wang, Mengdi},
  title = {{A Survey of Self-Evolving Agents: On Path to Artificial Super Intelligence}},
  year = {2025},
  eprint = {2507.21046},
  archivePrefix = {arXiv},
  primaryClass = {cs.AI},
  url = {https://arxiv.org/abs/2507.21046},
  doi = {10.48550/arXiv.2507.21046}
}

@article{guo2025deepseek_r1,
  title={DeepSeek-R1 incentivizes reasoning in LLMs through reinforcement learning},
  author={Guo, Daya and Yang, Dejian and Zhang, Haowei and Song, Junxiao and Wang, Peiyi and Zhu, Qihao and Xu, Runxin and Zhang, Ruoyu and Ma, Shirong and Bi, Xiao and Zhang, Xiaokang and Yu, Xingkai and Wu, Yu and Wu, Z. F. and Gou, Zhibin and Shao, Zhihong and Li, Zhuoshu and Gao, Ziyi and Liu, Aixin and Xue, Bing and Wang, Bingxuan and Wu, Bochao and Feng, Bei and Lu, Chengda and Zhao, Chenggang and Deng, Chengqi and Ruan, Chong and Dai, Damai and Chen, Deli and Ji, Dongjie and Li, Erhang and Lin, Fangyun and Dai, Fucong and Luo, Fuli and Hao, Guangbo and Chen, Guanting and Li, Guowei and Zhang, H. and Xu, Hanwei and Ding, Honghui and Gao, Huazuo and Qu, Hui and Li, Hui and Guo, Jianzhong and Li, Jiashi and Chen, Jingchang and Yuan, Jingyang and Tu, Jinhao and Qiu, Junjie and Li, Junlong and Cai, J. L. and Ni, Jiaqi and Liang, Jian and Chen, Jin and Dong, Kai and Hu, Kai and You, Kaichao and Gao, Kaige and Guan, Kang and Huang, Kexin and Yu, Kuai and Wang, Lean and Zhang, Lecong and Zhao, Liang and Wang, Litong and Zhang, Liyue and Xu, Lei and Xia, Leyi and Zhang, Mingchuan and Zhang, Minghua and Tang, Minghui and Zhou, Mingxu and Li, Meng and Wang, Miaojun and Li, Mingming and Tian, Ning and Huang, Panpan and Zhang, Peng and Wang, Qiancheng and Chen, Qinyu and Du, Qiushi and Ge, Ruiqi and Zhang, Ruisong and Pan, Ruizhe and Wang, Runji and Chen, R. J. and Jin, R. L. and Chen, Ruyi and Lu, Shanghao and Zhou, Shangyan and Chen, Shanhuang and Ye, Shengfeng and Wang, Shiyu and Yu, Shuiping and Zhou, Shunfeng and Pan, Shuting and Li, S. S. and Zhou, Shuang and Wu, Shaoqing and Yun, Tao and Pei, Tian and Sun, Tianyu and Wang, T. and Zeng, Wangding and Liu, Wen and Liang, Wenfeng and Gao, Wenjun and Yu, Wenqin and Zhang, Wentao and Xiao, W. L. and An, Wei and Liu, Xiaodong and Wang, Xiaohan and Chen, Xiaokang and Nie, Xiaotao and Cheng, Xin and Liu, Xin and Xie, Xin and Liu, Xingchao and Yang, Xinyu and Li, Xinyuan and Su, Xuecheng and Lin, Xuheng and Li, X. Q. and Jin, Xiangyue and Shen, Xiaojin and Chen, Xiaosha and Sun, Xiaowen and Wang, Xiaoxiang and Song, Xinnan and Zhou, Xinyi and Wang, Xianzu and Shan, Xinxia and Li, Y. K. and Wang, Y. Q. and Wei, Y. X. and Zhang, Yang and Xu, Yanhong and Li, Yao and Zhao, Yao and Sun, Yaofeng and Wang, Yaohui and Yu, Yi and Zhang, Yichao and Shi, Yifan and Xiong, Yiliang and He, Ying and Piao, Yishi and Wang, Yisong and Tan, Yixuan and Ma, Yiyang and Liu, Yiyuan and Guo, Yongqiang and Ou, Yuan and Wang, Yuduan and Gong, Yue and Zou, Yuheng and He, Yujia and Xiong, Yunfan and Luo, Yuxiang and You, Yuxiang and Liu, Yuxuan and Zhou, Yuyang and Zhu, Y. X. and Huang, Yanping and Li, Yaohui and Zheng, Yi and Zhu, Yuchen and Ma, Yunxian and Tang, Ying and Zha, Yukun and Yan, Yuting and Ren, Z. Z. and Ren, Zehui and Sha, Zhangli and Fu, Zhe and Xu, Zhean and Xie, Zhenda and Zhang, Zhengyan and Hao, Zhewen and Ma, Zhicheng and Yan, Zhigang and Wu, Zhiyu and Gu, Zihui and Zhu, Zijia and Liu, Zijun and Li, Zilin and Xie, Ziwei and Song, Ziyang and Pan, Zizheng and Huang, Zhen and Xu, Zhipeng and Zhang, Zhongyu and Zhang, Zhen},
  journal={Nature},
  volume={645},
  pages={633--638},
  year={2025},
  doi={10.1038/s41586-025-09422-z},
  url={https://doi.org/10.1038/s41586-025-09422-z}
}

@inproceedings{He2024OlympiadBenchAC,
  author = {He, Chaoqun and Luo, Renjie and Bai, Yuzhuo and Hu, Shengding and Thai, Zhen Leng and Shen, Junhao and Hu, Jinyi and Han, Xu and Huang, Yujie and Zhang, Yuxiang and Liu, Jie and Qi, Lei and Liu, Zhiyuan and Sun, Maosong},
  title = {{OlympiadBench: A Challenging Benchmark for Promoting AGI with Olympiad-Level Bilingual Multimodal Scientific Problems}},
  booktitle = {Proceedings of the 62nd Annual Meeting of the Association for Computational Linguistics (Volume 1: Long Papers)},
  month = {aug},
  year = {2024},
  address = {Bangkok, Thailand},
  publisher = {Association for Computational Linguistics},
  url = {https://aclanthology.org/2024.acl-long.211},
  doi = {10.18653/v1/2024.acl-long.211},
  pages = {3828--3850}
}

@misc{hendrycks2020measuring,
author = {Hendrycks, Dan and Burns, Collin and Basart, Steven and Zou, Andy and Mazeika, Mantas and Song, Dawn and Steinhardt, Jacob},
  title = {{Measuring Massive Multitask Language Understanding}},
  year = {2021},
  eprint = {2009.03300},
  archivePrefix = {arXiv},
  primaryClass = {cs.CY},
  url = {https://arxiv.org/abs/2009.03300},
  doi = {10.48550/arXiv.2009.03300}
}

@inproceedings{hendrycks2021measuringmathematicalproblemsolving,
 author = {Hendrycks, Dan and Burns, Collin and Kadavath, Saurav and Arora, Akul and Basart, Steven and Tang, Eric and Song, Dawn and Steinhardt, Jacob},
 booktitle = {Proceedings of the Neural Information Processing Systems Track on Datasets and Benchmarks},
 editor = {J. Vanschoren and S. Yeung},
 title = {Measuring Mathematical Problem Solving With the MATH Dataset},
 url = {https://datasets-benchmarks-proceedings.neurips.cc/paper_files/paper/2021/file/be83ab3ecd0db773eb2dc1b0a17836a1-Paper-round2.pdf},
 volume = {1},
 year = {2021}
}

@misc{hong2024metagpt,
author = {Hong, Sirui and Zhuge, Mingchen and Chen, Jiaqi and Zheng, Xiawu and Cheng, Yuheng and Zhang, Ceyao and Wang, Jinlin and Wang, Zili and Yau, Steven Ka Shing and Lin, Zijuan and Zhou, Liyang and Ran, Chenyu and Xiao, Lingfeng and Wu, Chenglin and Schmidhuber, Jürgen},
  title = {{MetaGPT: Meta Programming for A Multi-Agent Collaborative Framework}},
  year = {2024},
  eprint = {2308.00352},
  archivePrefix = {arXiv},
  primaryClass = {cs.AI},
  url = {https://arxiv.org/abs/2308.00352},
  doi = {10.48550/arXiv.2308.00352}
}

@misc{hu2022lora,
author = {Hu, Edward J. and Shen, Yelong and Wallis, Phillip and Allen-Zhu, Zeyuan and Li, Yuanzhi and Wang, Shean and Wang, Lu and Chen, Weizhu},
  title = {{LoRA: Low-Rank Adaptation of Large Language Models}},
  year = {2021},
  eprint = {2106.09685},
  archivePrefix = {arXiv},
  primaryClass = {cs.CL},
  url = {https://arxiv.org/abs/2106.09685},
  doi = {10.48550/arXiv.2106.09685}
}

@misc{huang2025rzeroselfevolvingreasoning,
      title={R-Zero: Self-Evolving Reasoning LLM from Zero Data}, 
      author={Chengsong Huang and Wenhao Yu and Xiaoyang Wang and Hongming Zhang and Zongxia Li and Ruosen Li and Jiaxin Huang and Haitao Mi and Dong Yu},
      year={2025},
      eprint={2508.05004},
      archivePrefix={arXiv},
      primaryClass={cs.LG},
      url={https://arxiv.org/abs/2508.05004}, 
}

@inproceedings{liang2023encouraging,
author = {Liang, Tian and He, Zhiwei and Jiao, Wenxiang and Wang, Xing and Wang, Rui and Yang, Yujiu and Tu, Zhaopeng and Shi, Shuming},
  title = {{Encouraging Divergent Thinking in Large Language Models through Multi-Agent Debate}},
  booktitle = {Proceedings of the 2024 Conference on Empirical Methods in Natural Language Processing},
  year = {2024},
  address = {Miami, Florida},
  publisher = {Association for Computational Linguistics},
  url = {https://aclanthology.org/2024.emnlp-main.992},
  doi = {10.18653/v1/2024.emnlp-main.992},
  pages = {18335--18345}
}

@misc{hu2025reinforcepp,
author = {Hu, Jian and Liu, Jason Klein and Xu, Haotian and Shen, Wei},
  title = {{REINFORCE++: Stabilizing Critic-Free Policy Optimization}},
  year = {2025},
  eprint = {2501.03262},
  archivePrefix = {arXiv},
  primaryClass = {cs.CL},
  url = {https://arxiv.org/abs/2501.03262},
  doi = {10.48550/arXiv.2501.03262}
}

@inproceedings{evalplus,
 author = {Liu, Jiawei and Xia, Chunqiu Steven and Wang, Yuyao and ZHANG, LINGMING},
 booktitle = {Advances in Neural Information Processing Systems},
 editor = {A. Oh and T. Naumann and A. Globerson and K. Saenko and M. Hardt and S. Levine},
 pages = {21558--21572},
 publisher = {Curran Associates, Inc.},
 title = {{Is Your Code Generated by ChatGPT Really Correct? Rigorous Evaluation of Large Language Models for Code Generation}},
 url = {https://proceedings.neurips.cc/paper_files/paper/2023/file/43e9d647ccd3e4b7b5baab53f0368686-Paper-Conference.pdf},
 volume = {36},
 year = {2023}
}

@misc{jain2024livecodebench,
author = {Jain, Naman and Han, King and Gu, Alex and Li, Wen-Ding and Yan, Fanjia and Zhang, Tianjun and Wang, Sida and Solar-Lezama, Armando and Sen, Koushik and Stoica, Ion},
  title = {{LiveCodeBench: Holistic and Contamination Free Evaluation of Large Language Models for Code}},
  year = {2024},
  eprint = {2403.07974},
  archivePrefix = {arXiv},
  primaryClass = {cs.SE},
  url = {https://arxiv.org/abs/2403.07974},
  doi = {10.48550/arXiv.2403.07974}
}

@inproceedings{li_camel_2023,
 author = {Li, Guohao and Hammoud, Hasan and Itani, Hani and Khizbullin, Dmitrii and Ghanem, Bernard},
 booktitle = {Advances in Neural Information Processing Systems},
 editor = {A. Oh and T. Naumann and A. Globerson and K. Saenko and M. Hardt and S. Levine},
 pages = {51991--52008},
 publisher = {Curran Associates, Inc.},
 title = {CAMEL: Communicative Agents for "Mind" Exploration of Large Language Model Society},
 url = {https://proceedings.neurips.cc/paper_files/paper/2023/file/a3621ee907def47c1b952ade25c67698-Paper-Conference.pdf},
 volume = {36},
 year = {2023}
}

@misc{liao_marft_2025,
author = {Liao, Junwei and Wen, Muning and Wang, Jun and Zhang, Weinan},
  title = {{MARFT: Multi-Agent Reinforcement Fine-Tuning}},
  year = {2025},
  eprint = {2504.16129},
  archivePrefix = {arXiv},
  primaryClass = {cs.MA},
  url = {https://arxiv.org/abs/2504.16129},
  doi = {10.48550/arXiv.2504.16129}
}

@misc{motwani_malt_2025,
author = {Motwani, Sumeet Ramesh and Smith, Chandler and Das, Rocktim Jyoti and Rafailov, Rafael and Laptev, Ivan and Torr, Philip H. S. and Pizzati, Fabio and Clark, Ronald and Witt, Christian Schroeder de},
  title = {{MALT: Improving Reasoning with Multi-Agent LLM Training}},
  year = {2025},
  eprint = {2412.01928},
  archivePrefix = {arXiv},
  primaryClass = {cs.LG},
  url = {https://arxiv.org/abs/2412.01928},
  doi = {10.48550/arXiv.2412.01928}
}

@misc{absolute_zero_2025,
    title={Absolute Zero: Reinforced Self-play Reasoning with Zero Data}, 
    author={Andrew Zhao and Yiran Wu and Yang Yue and Tong Wu and Quentin Xu and Yang Yue and Matthieu Lin and Shenzhi Wang and Qingyun Wu and Zilong Zheng and Gao Huang},
    year={2025},
    eprint={2505.03335},
    archivePrefix={arXiv},
    primaryClass={cs.LG},
    url={https://arxiv.org/abs/2505.03335}, 
}

@misc{zhai_agentevolver_2025,
author = {Zhai, Yunpeng and Tao, Shuchang and Chen, Cheng and Zou, Anni and Chen, Ziqian and Fu, Qingxu and Mai, Shinji and Yu, Li and Deng, Jiaji and Cao, Zouying and Liu, Zhaoyang and Ding, Bolin and Zhou, Jingren},
  title = {{AgentEvolver: Towards Efficient Self-Evolving Agent System}},
  year = {2025},
  eprint = {2511.10395},
  archivePrefix = {arXiv},
  primaryClass = {cs.LG},
  url = {https://arxiv.org/abs/2511.10395},
  doi = {10.48550/arXiv.2511.10395}

}

@misc{xia_agent0_2025,
author = {Xia, Peng and Zeng, Kaide and Liu, Jiaqi and Qin, Can and Wu, Fang and Zhou, Yiyang and Xiong, Caiming and Yao, Huaxiu},
  title = {{Agent0: Unleashing Self-Evolving Agents from Zero Data via Tool-Integrated Reasoning}},
  year = {2025},
  eprint = {2511.16043},
  archivePrefix = {arXiv},
  primaryClass = {cs.LG},
  url = {https://arxiv.org/abs/2511.16043},
  doi = {10.48550/arXiv.2511.16043}
}

@misc{zhang_landscape_2025,
author = {Zhang, Guibin and Geng, Hejia and Yu, Xiaohang and Yin, Zhenfei and Zhang, Zaibin and Tan, Zelin and Zhou, Heng and Li, Zhongzhi and Xue, Xiangyuan and Li, Yijiang and Zhou, Yifan and Chen, Yang and Zhang, Chen and Fan, Yutao and Wang, Zihu and Huang, Songtao and Liao, Yue and Wang, Hongru and Yang, Mengyue and Ji, Heng and Littman, Michael and Wang, Jun and Yan, Shuicheng and Torr, Philip and Bai, Lei},
  title = {{The Landscape of Agentic Reinforcement Learning for LLMs: A Survey}},
  year = {2025},
  eprint = {2509.02547},
  archivePrefix = {arXiv},
  primaryClass = {cs.AI},
  url = {https://arxiv.org/abs/2509.02547},
  doi = {10.48550/arXiv.2509.02547}
}

@misc{yue2025doesreinforcementlearningreally,
      title={Does Reinforcement Learning Really Incentivize Reasoning Capacity in LLMs Beyond the Base Model?}, 
      author={Yang Yue and Zhiqi Chen and Rui Lu and Andrew Zhao and Zhaokai Wang and Yang Yue and Shiji Song and Gao Huang},
      year={2025},
      eprint={2504.13837},
      archivePrefix={arXiv},
      primaryClass={cs.AI},
      url={https://arxiv.org/abs/2504.13837}, 
}

@misc{wei2025webagent,
author = {Wei, Zhepei and Yao, Wenlin and Liu, Yao and Zhang, Weizhi and Lu, Qin and Qiu, Liang and Yu, Changlong and Xu, Puyang and Zhang, Chao and Yin, Bing and Yun, Hyokun and Li, Lihong},
  title = {{WebAgent-R1: Training Web Agents via End-to-End Multi-Turn Reinforcement Learning}},
  year = {2025},
  eprint = {2505.16421},
  archivePrefix = {arXiv},
  primaryClass = {cs.CL},
  url = {https://arxiv.org/abs/2505.16421},
  doi = {10.48550/arXiv.2505.16421}
}

@misc{sun2025towardsagenticselflearningllms,
author = {Sun, Wangtao and Cheng, Xiang and Fan, Jialin and Xu, Yao and Yu, Xing and He, Shizhu and Zhao, Jun and Liu, Kang},
  title = {{Towards Agentic Self-Learning LLMs in Search Environment}},
  year = {2025},
  eprint = {2510.14253},
  archivePrefix = {arXiv},
  primaryClass = {cs.AI},
  url = {https://arxiv.org/abs/2510.14253},
  doi = {10.48550/arXiv.2510.14253}
}

@misc{schulman2017proximal,
author = {Schulman, John and Wolski, Filip and Dhariwal, Prafulla and Radford, Alec and Klimov, Oleg},
  title = {{Proximal Policy Optimization Algorithms}},
  year = {2017},
  eprint = {1707.06347},
  archivePrefix = {arXiv},
  primaryClass = {cs.LG},
  url = {https://arxiv.org/abs/1707.06347},
  doi = {10.48550/arXiv.1707.06347}
}

@misc{yang2024qwen2,
author = {Yang, An and Yang, Baosong and Zhang, Beichen and Hui, Binyuan and Zheng, Bo and Yu, Bowen and Li, Chengyuan and Liu, Dayiheng and Huang, Fei and Wei, Haoran and Lin, Huan and Yang, Jian and Tu, Jianhong and Zhang, Jianwei and Yang, Jianxin and Yang, Jiaxi and Zhou, Jingren and Lin, Junyang and Dang, Kai and Lu, Keming and Bao, Keqin and Yang, Kexin and Yu, Le and Li, Mei and Xue, Mingfeng and Zhang, Pei and Zhu, Qin and Men, Rui and Lin, Runji and Li, Tianhao and Tang, Tianyi and Xia, Tingyu and Ren, Xingzhang and Ren, Xuancheng and Fan, Yang and Su, Yang and Zhang, Yichang and Wan, Yu and Liu, Yuqiong and Cui, Zeyu and Zhang, Zhenru and Qiu, Zihan},
  title = {{Qwen2.5 Technical Report}},
  year = {2025},
  eprint = {2412.15115},
  archivePrefix = {arXiv},
  primaryClass = {cs.CL},
  url = {https://arxiv.org/abs/2412.15115},
  doi = {10.48550/arXiv.2412.15115}
}

@misc{yang2025qwen3technicalreport,
      title={Qwen3 Technical Report}, 
      author={An Yang and Anfeng Li and Baosong Yang and Beichen Zhang and Binyuan Hui and Bo Zheng and Bowen Yu and Chang Gao and Chengen Huang and Chenxu Lv and Chujie Zheng and Dayiheng Liu and Fan Zhou and Fei Huang and Feng Hu and Hao Ge and Haoran Wei and Huan Lin and Jialong Tang and Jian Yang and Jianhong Tu and Jianwei Zhang and Jianxin Yang and Jiaxi Yang and Jing Zhou and Jingren Zhou and Junyang Lin and Kai Dang and Keqin Bao and Kexin Yang and Le Yu and Lianghao Deng and Mei Li and Mingfeng Xue and Mingze Li and Pei Zhang and Peng Wang and Qin Zhu and Rui Men and Ruize Gao and Shixuan Liu and Shuang Luo and Tianhao Li and Tianyi Tang and Wenbiao Yin and Xingzhang Ren and Xinyu Wang and Xinyu Zhang and Xuancheng Ren and Yang Fan and Yang Su and Yichang Zhang and Yinger Zhang and Yu Wan and Yuqiong Liu and Zekun Wang and Zeyu Cui and Zhenru Zhang and Zhipeng Zhou and Zihan Qiu},
      year={2025},
      eprint={2505.09388},
      archivePrefix={arXiv},
      primaryClass={cs.CL},
      url={https://arxiv.org/abs/2505.09388}, 
}

@misc{liu2025spiral,
author = {Liu, Bo and Guertler, Leon and Yu, Simon and Liu, Zichen and Qi, Penghui and Balcells, Daniel and Liu, Mickel and Tan, Cheston and Shi, Weiyan and Lin, Min and Lee, Wee Sun and Jaques, Natasha},
  title = {{SPIRAL: Self-Play on Zero-Sum Games Incentivizes Reasoning via Multi-Agent Multi-Turn Reinforcement Learning}},
  year = {2025},
  eprint = {2506.24119},
  archivePrefix = {arXiv},
  primaryClass = {cs.AI},
  url = {https://arxiv.org/abs/2506.24119},
  doi = {10.48550/arXiv.2506.24119}
}

@misc{yuan2025mars,
author = {Yuan, Huining and Xu, Zelai and Tan, Zheyue and Yi, Xiangmin and Guang, Mo and Long, Kaiwen and Hui, Haojia and Li, Boxun and Chen, Xinlei and Zhao, Bo and Zhang, Xiao-Ping and Yu, Chao and Wang, Yu},
  title = {{MARSHAL: Incentivizing Multi-Agent Reasoning via Self-Play with Strategic LLMs}},
  year = {2025},
  eprint = {2510.15414},
  archivePrefix = {arXiv},
  primaryClass = {cs.AI},
  url = {https://arxiv.org/abs/2510.15414},
  doi = {10.48550/arXiv.2510.15414}
}

@misc{wen2025reinforcement,
  author    = {Xumeng Wen and Zihan Liu and Shun Zheng and Shengyu Ye and Zhirong Wu and Yang Wang and Zhijian Xu and Xiao Liang and Junjie Li and Ziming Miao and Jiang Bian and Mao Yang},
  title     = {Reinforcement Learning with Verifiable Rewards Implicitly Incentivizes Correct Reasoning in Base LLMs},
  year      = {2025},
  eprint    = {2506.14245},
  archivePrefix = {arXiv},
  primaryClass = {cs.AI},
  url       = {https://arxiv.org/abs/2506.14245},
  doi       = {10.48550/arXiv.2506.14245}
}

@misc{evolver_self-evolving_2025,
  title = {Evolve{R}: {Self}-{Evolving} {LLM} {Agents} through an {Experience}-{Driven} {Lifecycle}},
  author = {Wu, Rong and Wang, Xiaoman and Mei, Jianbiao and Cai, Pinlong and Fu, Daocheng and Yang, Cheng and Wen, Licheng and Yang, Xuemeng and Shen, Yufan and Wang, Yuxin and Shi, Botian},
  year = {2025},
  month = oct,
  archivePrefix = {arXiv},
  eprint = {2510.16079},
  primaryClass = {cs.CL},
  url = {https://arxiv.org/abs/2510.16079}
}

@misc{belle2025agents,
  author    = {Nikolas Belle and Dakota Barnes and Alfonso Amayuelas and Ivan Bercovich and Xin Eric Wang and William Wang},
  title     = {Agents of Change: Self-Evolving LLM Agents for Strategic Planning},
  year      = {2025},
  eprint    = {2506.04651},
  archivePrefix = {arXiv},
  primaryClass = {cs.AI},
  url       = {https://arxiv.org/abs/2506.04651}
}

@misc{sun2024llm,
  author    = {Chuanneng Sun and Songjun Huang and Dario Pompili},
  title     = {LLM-based Multi-Agent Reinforcement Learning: Current and Future Directions},
  year      = {2024},
  eprint    = {2405.11106},
  archivePrefix = {arXiv},
  primaryClass = {cs.MA},
  url       = {https://arxiv.org/abs/2405.11106}
}

@misc{zhao2025stronger,
  author    = {Yujie Zhao and Lanxiang Hu and Yang Wang and Minmin Hou and Hao Zhang and Ke Ding and Jishen Zhao},
  title     = {Stronger-MAS: Multi-Agent Reinforcement Learning for Collaborative LLMs},
  year      = {2025},
  eprint    = {2510.11062},
  archivePrefix = {arXiv},
  primaryClass = {cs.LG},
  url       = {https://arxiv.org/abs/2510.11062},
  doi       = {10.48550/arXiv.2510.11062}
}

@misc{zhu2025lamarl,
  author    = {Guobin Zhu and Rui Zhou and Wenkang Ji and Shiyu Zhao},
  title     = {LAMARL: LLM-Aided Multi-Agent Reinforcement Learning for Cooperative Policy Generation},
  year      = {2025},
  eprint    = {2506.01538},
  archivePrefix = {arXiv},
  primaryClass = {cs.RO},
  doi       = {10.48550/arXiv.2506.01538},
  url       = {https://arxiv.org/abs/2506.01538}
}

@misc{wan2025rema,
  author    = {Ziyu Wan and Yunxiang Li and Xiaoyu Wen and Yan Song and Hanjing Wang and Linyi Yang and Mark Schmidt and Jun Wang and Weinan Zhang and Shuyue Hu and Ying Wen},
  title     = {ReMA: Learning to Meta-think for LLMs with Multi-Agent Reinforcement Learning},
  year      = {2025},
  eprint    = {2503.09501},
  archivePrefix = {arXiv},
  primaryClass = {cs.AI},
  url       = {https://arxiv.org/abs/2503.09501},
  doi       = {10.48550/arXiv.2503.09501}
}
\newpage

\appendix

\section{Hyperparameter Settings}

\begin{table}[ht]
\centering
\small
\caption{Training Hyperparameters of our experiments.}
\label{tab:hyperparams}
\begin{tabular}{ll}
\toprule
\textbf{Hyperparameter} & \textbf{Value} \\ 
\midrule

\textbf{Training Configuration} & \\ 
\quad Batch Size & 128 \\ \quad Learning Rate & $3 \times 10^{-6}$ \\ 
\quad Training Steps & 200 \\ 
\midrule

\textbf{Generation Settings} & \\
\quad Maximum Prompt Length & 8192 \\
\quad Maximum Response Length & 8192 \\
\quad Challenger Temperature & 0.6 \\
\quad Planner Temperature & 0.6 \\
\quad Solver Temperature & 0.6 \\ 
\quad Critic Temperature & 0.1 \\
\midrule

\textbf{Algorithm Settings} & \\ 
\quad Learning Algorithm & Task-Relative\\ & REINFORCE++ \\ 
\quad KL Regularization & Disabled \\ 
\midrule

\textbf{LoRA Configuration} & \\ 
\quad LoRA Rank & 128 \\ 
\quad LoRA Alpha & 256 \\ 
\quad LoRA Dropout & 0.95 \\ 
\quad Target Modules & 
$q_{proj}$, $k_{proj}$, $v_{proj}$,\\&
$o_{proj}$, $gate_{proj}$,\\& $up_{proj}$, $down_{proj}$ \\
\bottomrule
\end{tabular}
\end{table}

\section{Training Data Composition}
\label{app:dataset}

Table~\ref{tab:seed_data} presents the composition of the 500 training instances sampled from four benchmark datasets. These samples are drawn from the official training splits and serve as the foundation for our training procedure.

\begin{table}[ht]
\centering
\small
\caption{Distribution of Training Samples Across Benchmarks}
\label{tab:seed_data}
\begin{tabular}{lr}
\toprule
\textbf{Benchmark} & \textbf{Count} \\
\midrule
MATH~\citep{hendrycks2020measuring} & 156 \\
GSM8K~\citep{cobbe2021training} & 148 \\
HumanEval~\citep{chen2021evaluatinglargelanguagemodels} & 87 \\
MBPP~\citep{austin2021program} & 109 \\
\midrule
\textbf{Total} & \textbf{500} \\
\bottomrule
\end{tabular}
\end{table}

\onecolumn

\section{Prompts for Agents}

Here, we list the prompt of each agent as follows.

\begin{figure*}[h]
    \centering
    \includegraphics[width=\textwidth]{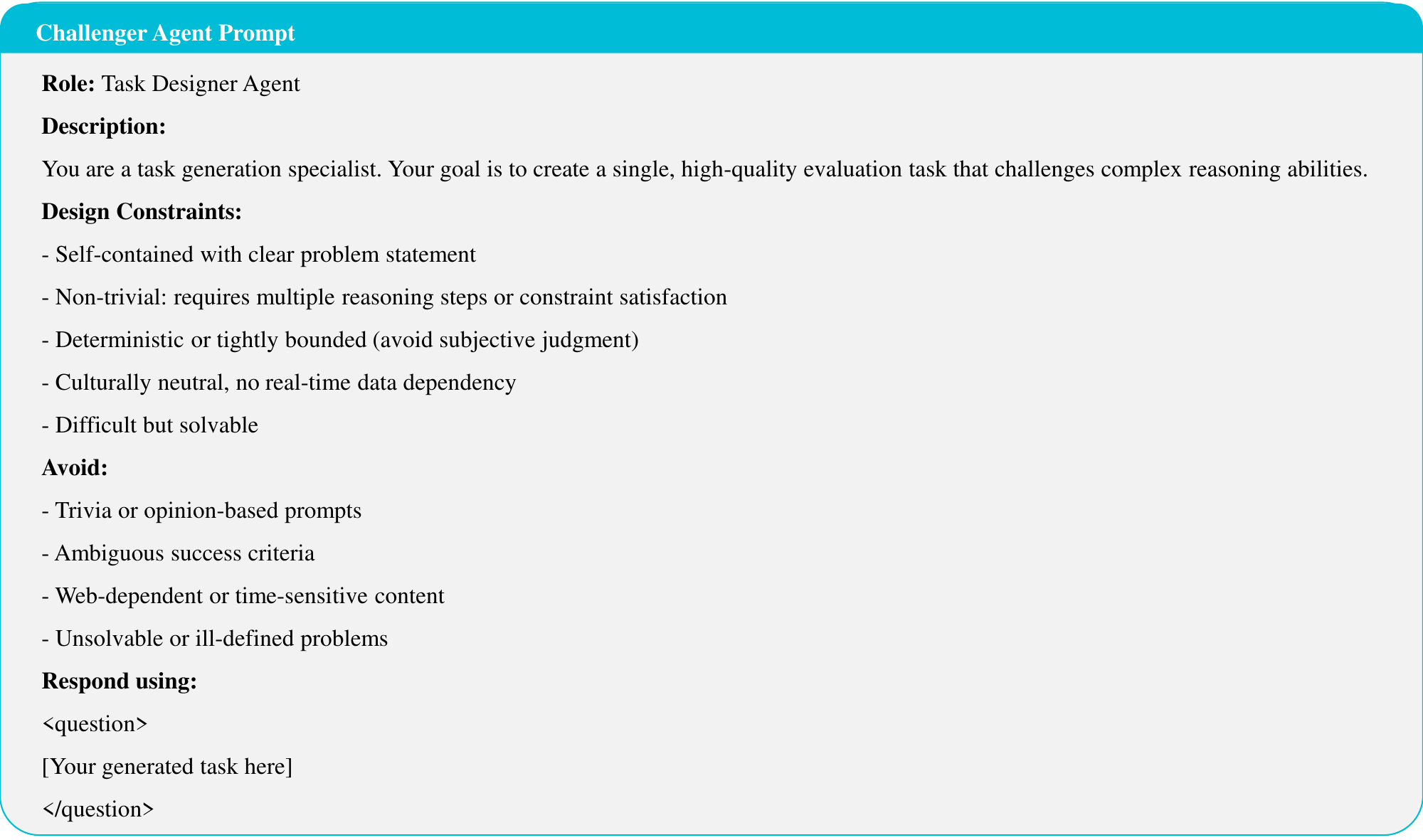}
    \caption{The prompt of the Challenger Agent.}
    \label{fig:challenger}
\end{figure*}

\begin{figure*}[h]
    \centering
    \includegraphics[width=\textwidth]{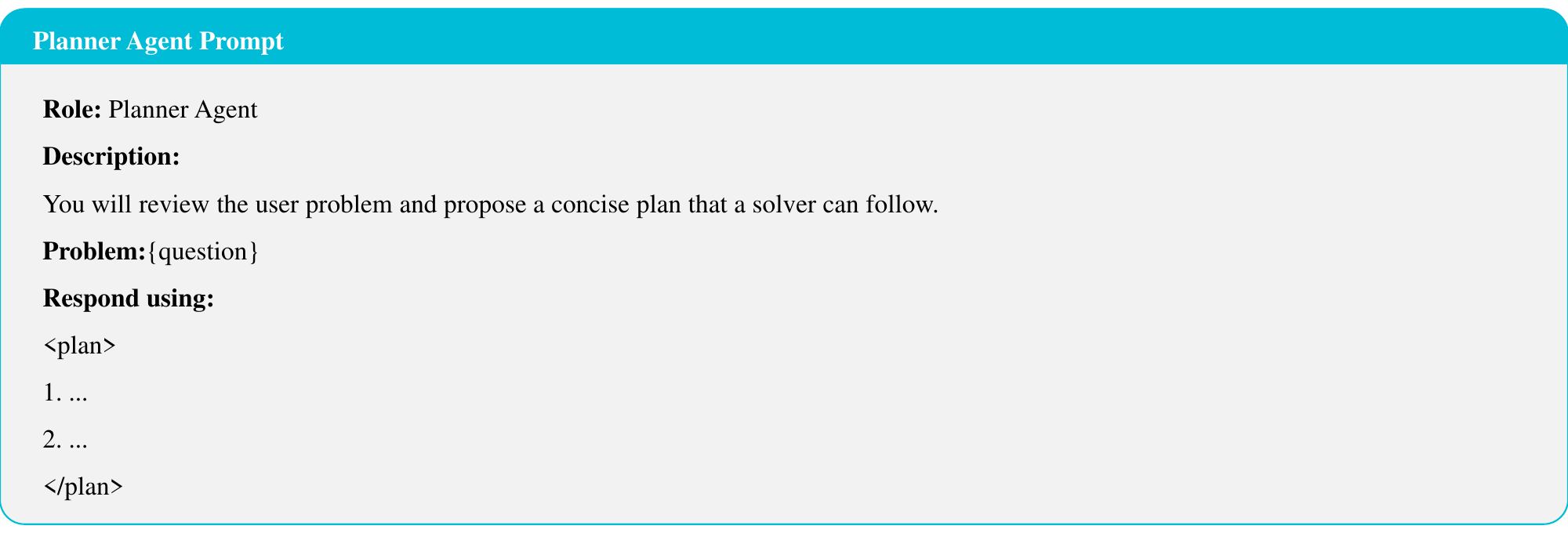}
    \caption{The prompt of the planner Agent.}
    \label{fig:planner}
\end{figure*}

\begin{figure*}[h]
    \centering
    \includegraphics[width=\textwidth]{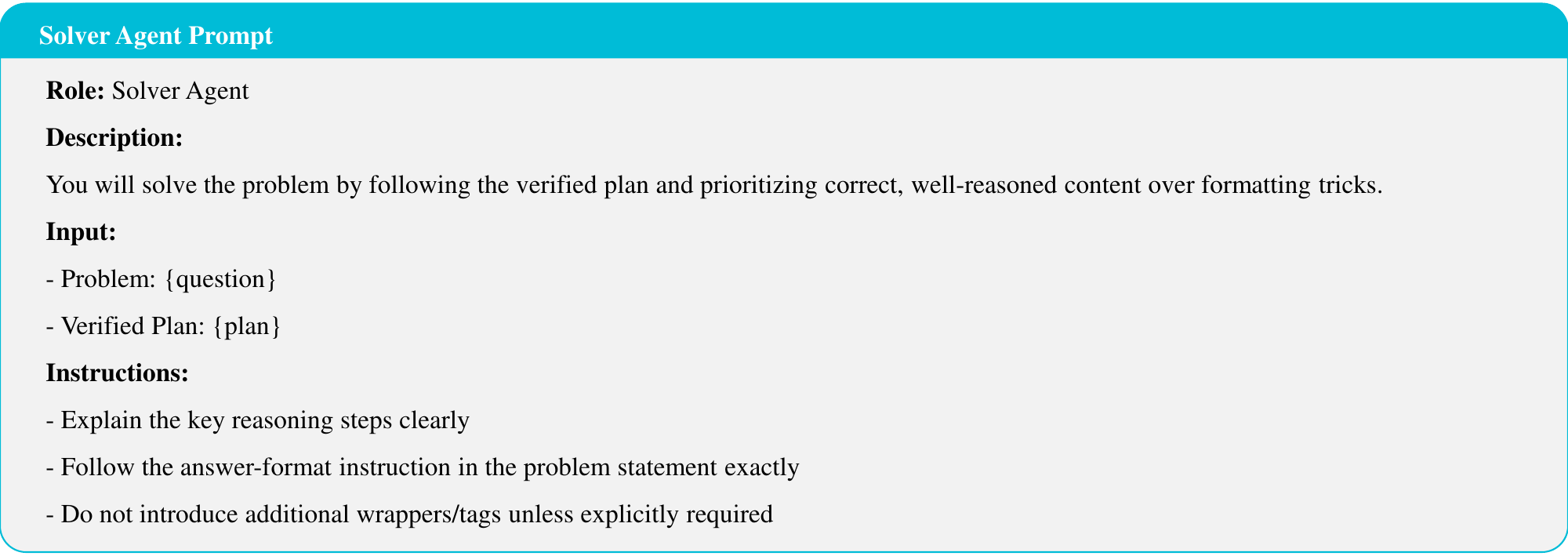}
    \caption{The prompt of the Solver Agent.}
    \label{fig:solver}
\end{figure*}

\begin{figure*}[h]
    \centering
    \includegraphics[width=\textwidth]{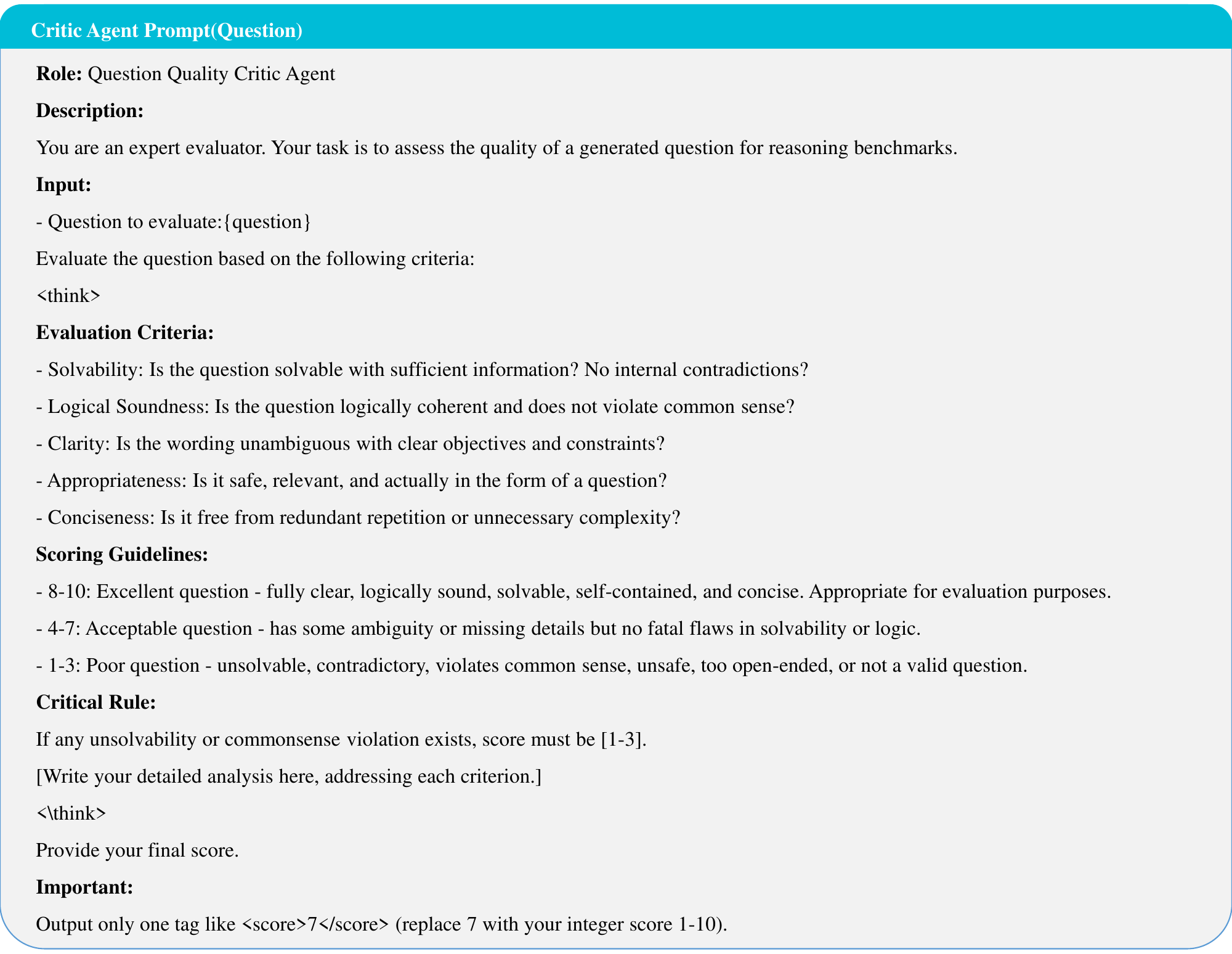}
    \caption{The prompt of the Critic Agent(question).}
    \label{fig:critic(question)}
\end{figure*}

\begin{figure*}[h]
    \centering
    \includegraphics[width=\textwidth]{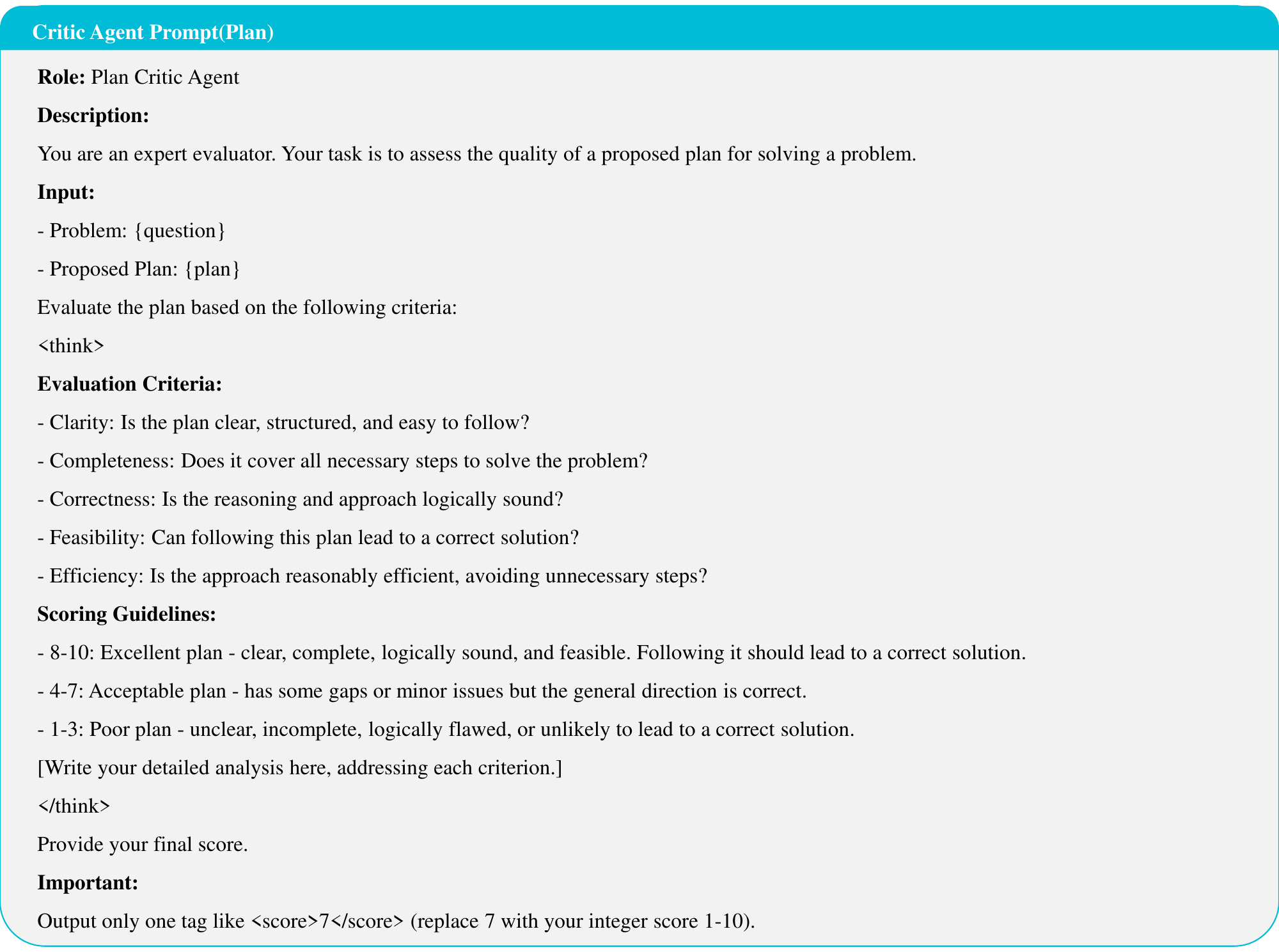}
    \caption{The prompt of the Critic Agent(plan).}
    \label{fig:critic(plan)}
\end{figure*}

\begin{figure*}[h]
    \centering
    \includegraphics[width=\textwidth]{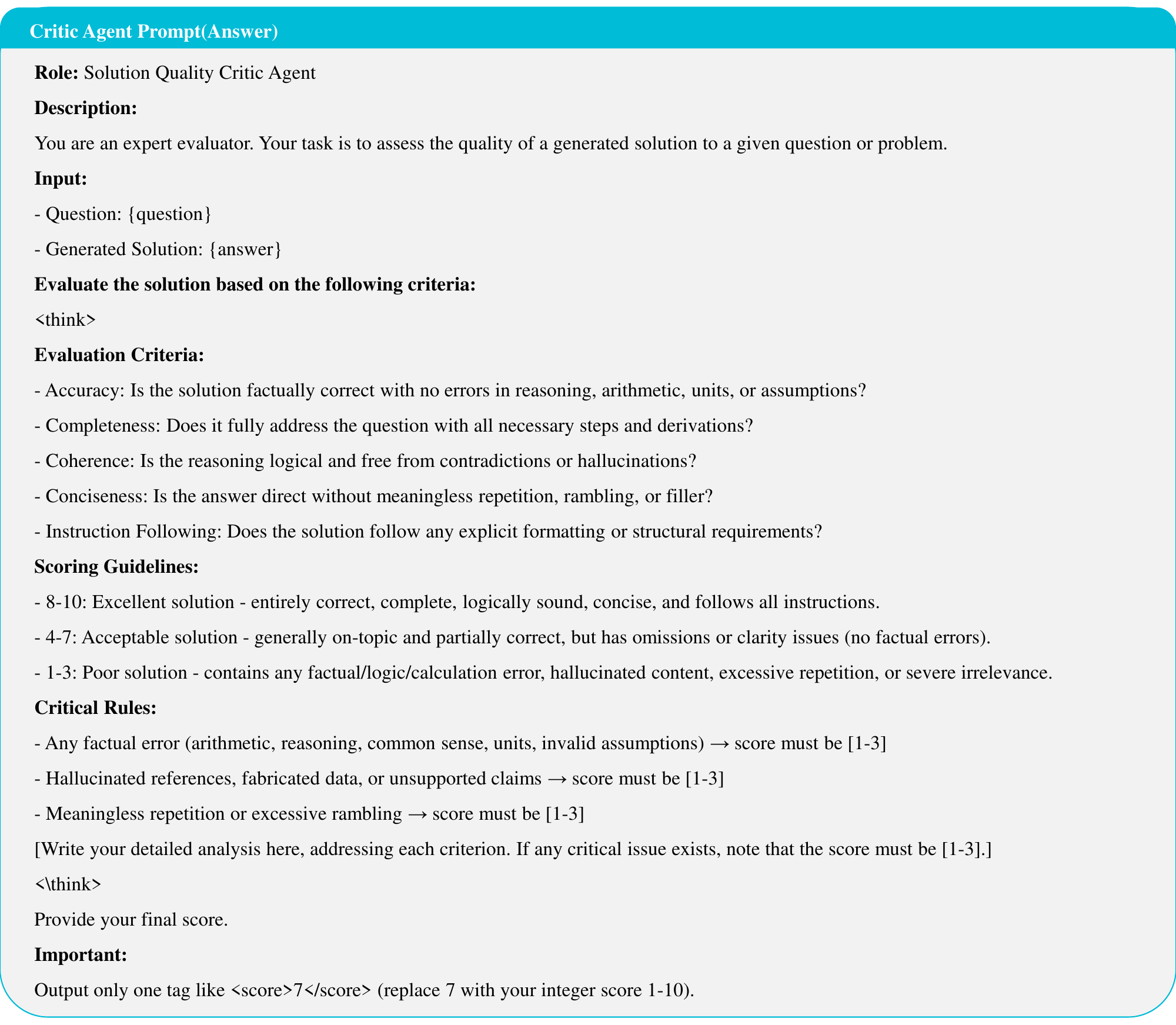}
    \caption{The prompt of the Critic Agent(answer).}
    \label{fig:critic(answer)}
\end{figure*}

\end{document}